\pdfoutput=1
\documentclass[11pt]{article}

\usepackage[a4paper, margin=2.5cm]{geometry}
\usepackage[T1]{fontenc}
\usepackage[super,comma,sort&compress]{natbib}
\usepackage{booktabs}
\usepackage{array}
\newcolumntype{L}[1]{>{\raggedright\arraybackslash}p{#1}}
\usepackage{graphicx}
\usepackage{caption}
\usepackage{xcolor}
\usepackage{amsmath}
\usepackage{amssymb}
\usepackage{authblk}
\usepackage{url}
\usepackage[scaled=0.92]{helvet}
\usepackage{sansmath}
\usepackage{tikz}
\usepackage{placeins}
\usetikzlibrary{arrows.meta,calc,decorations.pathreplacing}

\definecolor{hlPull}{RGB}{28,100,166}
\definecolor{hlPush}{RGB}{212,16,32}
\definecolor{hlConv}{RGB}{120,150,110}
\definecolor{hlFull}{RGB}{150,150,150}
\definecolor{hlTile}{RGB}{120,120,120}
\definecolor{hlTail}{RGB}{110,110,110}

\newtheorem{theorem}{Theorem}
\newtheorem{lemma}{Lemma}
\newtheorem{proposition}{Proposition}
\newcommand{\model}{\textsc{Astra}}
\newcommand{\hla}{Kantorovich Attention}

\newif\ifsubmission
\submissionfalse
\ifdefined\buildfigsatend\submissiontrue\fi

\newcounter{edtable}
\newcommand{\edtable}[1]{%
  \refstepcounter{edtable}%
  \par\medskip\noindent{\small\textbf{Extended Data Table
  \arabic{edtable}\,$\vert$\,#1}}\par\smallskip}
\newcounter{edfig}
\newcommand{\edfig}[1]{%
  \refstepcounter{edfig}%
  \par\medskip\noindent{\small\textbf{Extended Data Fig.\
  \arabic{edfig}\,$\vert$\,#1}}\par\smallskip}

\title{\bfseries\LARGE
Efficient probabilistic weather forecasting\\
at quarter-degree resolution}
\author[1]{Junu Choi}
\author[1,2]{Youngjoon Hong\thanks{Corresponding author.}}
\affil[1]{\small Interdisciplinary Program in Artificial Intelligence,
  Seoul National University, Seoul, Republic of Korea}
\affil[2]{\small Department of Mathematical Sciences,
  Seoul National University, Seoul, Republic of Korea}
\affil[ ]{\small Emails: \texttt{junu0413@snu.ac.kr} (J.C.),
  \texttt{hongyj@snu.ac.kr} (Y.H.)}
\date{}

\begin{document}
\maketitle

\section{Abstract}
Machine-learning ensembles deliver skilful 0.25\textdegree\ weather forecasts but remain costly to train and run.
Here we present \model, a global probabilistic model whose training cost is 9.8\% of GenCast's and whose 50-member, 15-day forecast takes 34 minutes on one consumer GPU using 19\,GiB of memory.
\model\ advances 75 variables by six hours in one network evaluation.
Within each step, Kantorovich Attention lets each region select its senders and receivers.
During training, a self-featurized continuous ranked probability score (CRPS) compares how a frozen copy of \model\ represents predicted and reanalysis states in later forecast inputs.
On WeatherBench~2, \model\ is better than ECMWF's ensemble on 98\% of 570 target and lead-time combinations and comparable to GenCast in both CRPS and ensemble-mean root-mean-square error.
At day 5, \model's mean tropical cyclone track error is 146\,km lower than that of ECMWF's ensemble.

\vspace{10mm}
\ifsubmission\else
\section{Introduction}
\fi

Machine learning has transformed medium-range weather forecasting
in recent years\cite{lam2023,bi2023,chen2023}.
Ensemble forecasting models generate multiple plausible weather trajectories to represent predictive uncertainty.
Models such as GenCast \cite{price2025} have surpassed ECMWF's operational ensemble, IFS ENS\cite{haiden2021}, on standard forecast  benchmarks\cite{alet2025}.
These models learn from decades of ERA5 reanalysis\cite{hersbach2020}, with several operating at 0.25\textdegree.
At this resolution, models can represent spatial detail in fronts,
tropical cyclones and terrain-related weather patterns that coarser
forecast grids smooth out.

Despite these advances, computational cost remains a major constraint on high-resolution ensemble forecasting.
Diffusion-based models such as GenCast require repeated network
evaluations to generate each forecast step\cite{price2025}, increasing the cost of producing large ensembles.
FGN and FourCastNet~3 cut this inference cost by generating each step in one network evaluation\cite{alet2025,bonev2025}. Both cost more to train than GenCast.
ArchesWeatherGen and U-Cast train at low cost\cite{couairon2026,cachay2026}, but on a 1.5\textdegree\ grid with 36 times fewer points than a 0.25\textdegree\ grid. Per grid point, their training costs more than GenCast's.
At 0.25\textdegree, each global field contains more than
a million grid points, making it costly to learn global interactions and to train on multi-step forecasts.
A key challenge therefore remains: retaining accurate 0.25\textdegree{} ensemble forecasts while reducing both training and inference costs.

To address this computational barrier, we developed \model, a global probabilistic weather forecasting model that combines accurate 0.25\textdegree{} forecasts of 75 atmospheric and surface variables with efficient training and fast inference.

\model\ and GenCast reach the same skill averaged over variables and lead times, while \model\ trains at 9.8\% of GenCast's cost.
\model\ generates each forecast step in a single network evaluation.
A 50-member, 15-day forecast takes 34 minutes within 19\,GiB of memory on one consumer GPU.

\model's design centres on two complementary advances: efficient global information exchange and a training objective that makes better use of short generated trajectories.

The key architectural component of \model\ is \hla\ (KA), which reduces the cost of global interactions within each training update while ideally reaching the expressive power of dense attention.
Atmospheric processes couple regions thousands of kilometres apart.
Dense attention\cite{vaswani2017} reaches the whole globe, but its cost is prohibitive on grids of more than a million points.
Local windows\cite{liu2021swin,bi2023} reach only nearby regions, and spherical filters\cite{bonev2023sfno,bonev2025} do not adapt to the atmospheric state.
Content-based sparse attention, as in BiFormer\cite{zhu2023biformer,yuan2025nsa,zhdanov2026}, is global and adaptive, but only the receivers choose their senders.
Selection on one side can leave tiles unselected and miss the best pairing of senders and receivers.
In KA, each tile, a block of grid points, selects both $K$ senders and $K$ receivers.
This guarantees that every tile sends to at least $K$ others, and the selected pairs can support minimum-cost transport plans in Kantorovich's formulation\cite{kantorovich1942}, which gives KA its name.

The key training component is self-featurized CRPS (SF-CRPS), a form of the continuous ranked probability score (CRPS) that improves forecasts days ahead using two generated steps.
In autoregressive forecasting, each predicted state becomes an input to subsequent steps.
Models that generate each step in one evaluation are trained with CRPS in fair or almost-fair (afCRPS) form\cite{pacchiardi2024,gneiting2007,lang2024,alet2025}.
Supervising later lead times then takes longer generated rollouts, each step adding a network evaluation and its backpropagation.
Applied to each grid point and variable separately, CRPS also misses how variables and locations relate.
SF-CRPS addresses both with the features of a frozen copy of the forecasting model itself, not a separately trained encoder.
Each feature combines several variables and neighbouring locations.
During training, \model\ generates forecasts at 6 and 12 h. For a later forecast step, the frozen copy processes an input window containing one or both predictions, with ERA5 supplying the remaining states. SF-CRPS scores the resulting ensemble of features against the features extracted from the corresponding ERA5 window.
In a quarter-width network, SF-CRPS lowers CRPS by 3.0\% at equal training time relative to afCRPS alone.

With these two components, \model\ achieves a lower CRPS than IFS ENS on 560 of 570 target and lead-time combinations of the WeatherBench~2 benchmark\cite{rasp2024}.
When our released forecasts were scored by WeatherBench~2, both CRPS and ensemble-mean root-mean-square error (RMSE) were lower than those of IFS ENS on all 600 combinations, and CRPS was lower than that of GenCast on 513.
From day 9, \model's ensemble spread matches its error more closely than those of GenCast and IFS ENS, and its members stay sharper than deterministic forecasts\cite{lam2023,chen2023,bi2023}.
\model\ also tracks tropical cyclones 146\,km more closely than IFS ENS at day 5, and its precipitation forecasts are significantly better at every lead time.

\section{Results}

\subsection{Forecast skill and computational cost}
\ifsubmission\else\ifsubmission
\begin{figure*}[!ht]
\else
\begin{figure*}[!t]
\fi
\centering
\includegraphics[width=\textwidth]{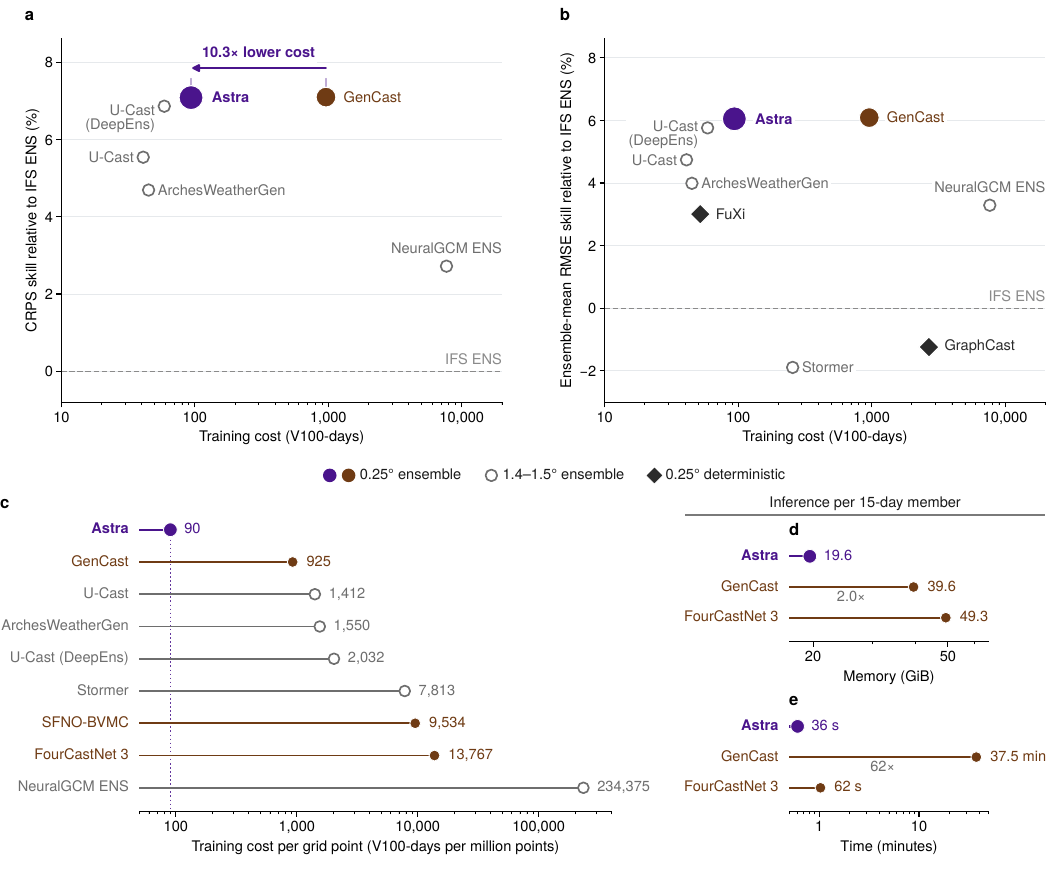}
\par\smallskip
\begin{minipage}{\textwidth}\footnotesize
\refstepcounter{figure}\label{fig:efficiency}%
\textbf{Fig.~\thefigure\ $\vert$ Competitive quarter-degree forecast skill at low training and inference cost.}
\textbf{a},~CRPS skill relative to IFS ENS against training cost. \textbf{b},~As \textbf{a}, for ensemble-mean RMSE skill.
Skill is the percentage by which a score is lower than that of IFS ENS, so positive values are better. Following U-Cast\cite{cachay2026}, skill is averaged over days 1--10 and over z500, q700, t850, u850 and v850
in \textbf{a} and over z500, t850, u850 and v850 in \textbf{b}. For the deterministic models GraphCast and FuXi, \textbf{b} uses the RMSE of the single forecast.
\model\ (50 members), GenCast\cite{price2025}, GraphCast\cite{lam2023} and FuXi\cite{chen2023} are scored
at 0.25\textdegree\ (1,038,240 grid points) on
732 initializations of 2020 (702 for FuXi). Other scores, including NeuralGCM ENS\cite{kochkov2024} and a stochastic version of Stormer\cite{nguyen2024},
are from U-Cast at 1.5\textdegree\ (29,040 grid points).
\textbf{c},~Training cost per grid point of the ensembles in \textbf{a} and \textbf{b} and of
SFNO-BVMC\cite{mahesh2025} and FourCastNet~3\cite{bonev2025}. The dotted line marks the value of \model.
\textbf{d},~Peak memory of one 15-day member at inference. \textbf{e},~Runtime of the same member.
All three models run on one RTX~PRO~6000.
Grey numbers in \textbf{d} and \textbf{e} are the ratios of GenCast's values to those of \model.
\end{minipage}
\end{figure*}

\fi%
On the WeatherBench~2 evaluation for 2020\cite{rasp2024}, \model\ achieves a 7.1\% lower CRPS and a 6.1\% lower ensemble-mean RMSE than IFS ENS, averaged over the variables and lead times of Fig.~\ref{fig:efficiency}a,b. CRPS scores the whole distribution of the ensemble against the ground truth, and the ensemble-mean RMSE scores the average of its members. GenCast\cite{price2025} achieves a 7.1\% lower CRPS and a 6.1\% lower ensemble-mean RMSE than IFS ENS, equal to those of \model. \model's released forecasts have been scored by WeatherBench~2, and the scores are listed alongside those of other models on its website, at \url{https://sites.research.google/gr/weatherbench/probabilistic-scores/}.

\model\ reaches this skill at a 10.3 times lower training cost than GenCast. Training cost is measured in V100-days, which convert the training days of each accelerator to those of a V100 GPU with the ratios of ArchesWeatherGen and U-Cast\cite{couairon2026,cachay2026}. The two training stages of \model\ take 56\,h on one node of eight H200 GPUs, or 93.6 V100-days. Figure~\ref{fig:efficiency}c divides the training cost by the number of grid points of each model. Per grid point, every other ensemble costs at least 10.3 times as much to train as \model.

At inference, a 15-day member of \model\ takes 36\,s and 19.6\,GiB, the least time and memory of the three models that Fig.~\ref{fig:efficiency}d,e compares on the same GPU. A member of FourCastNet~3 takes 62\,s and 49.3\,GiB, and a member of GenCast takes 37.5\,min and 39.6\,GiB. Of the three, only \model\ stays within the 32\,GiB of a single consumer GPU. On an RTX~5090, a 15-day member of \model\ takes 41\,s and 19.0\,GiB, and a 50-member forecast takes 34\,min.

\subsection{The model and an example forecast}
\ifsubmission\else\ifsubmission
\begin{figure}[!ht]
\else
\begin{figure}[!tp]
\fi
\centering
\includegraphics[width=\textwidth]{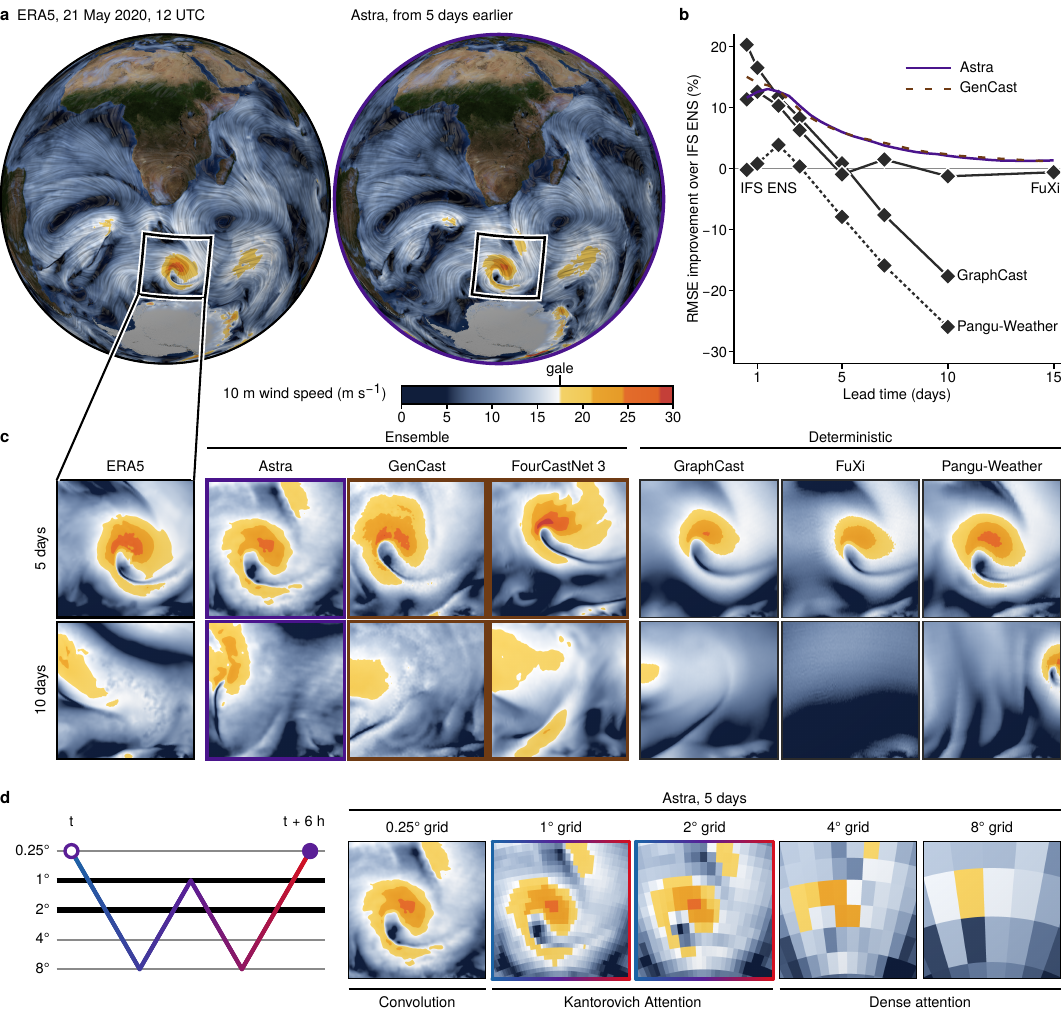}
\par\smallskip
\begin{minipage}{\textwidth}\footnotesize
\refstepcounter{figure}\label{fig:forecast}%
\textbf{Fig.~\thefigure\ \textbar\ \model\ produces sharp members and an ensemble mean more accurate than that of IFS ENS using a W-Net with two-sided transport attention.}
\textbf{a},~10\,m wind in ERA5 and in one \model\ member. Tone shows wind speed, turning yellow at gale force (17.5\,m\,s$^{-1}$), and streaks show direction; the
square marks the storm shown in \textbf{c} and \textbf{d}.
\textbf{b},~Median RMSE improvement over the IFS ENS ensemble mean across 12 WeatherBench~2 targets (z500, t500, t850, u500, u850, v500, v850, t2m, msl, u10, v10, ws10; abbreviations as in Fig.~\ref{fig:scorecard}) in 2020, for the
ensemble means of \model\ and GenCast\cite{price2025} and the single forecasts of GraphCast\cite{lam2023},
FuXi\cite{chen2023} and Pangu-Weather\cite{bi2023}, which are marked by diamonds.
\textbf{c},~10\,m wind speed in the square in ERA5, in one member of each ensemble and in the deterministic
forecasts. The rows show 5 and 10 days after 12\,UTC on 16 May 2020.
\textbf{d},~One network evaluation takes the state from $t$ to $t+6$\,h through the five grids of the W-Net.
Heavy strokes mark \hla\ (KA). The blue-to-red gradient marks \model's network and the frames mark the grids on which KA acts,
in the pull and push colours of Fig.~\ref{fig:ka}. The squares show \model's 5-day wind of \textbf{c} averaged to each grid.
\end{minipage}
\end{figure}

\fi%
From the four latest states at 6\,h intervals and a random noise field, \model\ generates the next state in one network
evaluation. A state comprises five atmospheric fields on 13 pressure levels, four surface fields and six auxiliary fields on the 0.25\textdegree{} grid of ERA5, and a forecast runs for 15~days in 60 steps. \model's network is a W-Net: two U-Nets applied in sequence, so that the
state in Fig.~\ref{fig:forecast}d passes down and up through five resolution levels twice. The W-Net uses
convolutions on the finest level, KA at 1\textdegree\ and 2\textdegree\ and dense attention on the two coarsest. The network is trained on ERA5 of 1979--2019, with details in Extended Data Fig.~\ref{ed:arch} and Extended
Data Table~\ref{edt:cost}. The noise has the spatial and cross-variable structure
of ERA5's 6\,h residuals, the changes of the state over 6\,h, as in Extended Data Fig.~\ref{ed:prior}. A single network is trained from scratch in two stages, without a pretrained deterministic model or distillation.
The first stage uses MeanFlow\cite{geng2025}, which trains the network to turn a random noise field into a sample of the 6\,h residual in one evaluation.
The second stage uses SF-CRPS on forecasts of two steps. Because every step makes fresh random draws, members from the same
initial states differ.

Ensemble members remain sharp where deterministic forecasts smooth, as in a deep storm over the Southern Ocean.
One \model\ member is set alongside ERA5 in Fig.~\ref{fig:forecast}a for 21 May 2020, five days ahead, with the storm
boxed on both globes. At that lead time, the storm's ensemble-mean centre lies 91\,km from that in ERA5 for
\model, against 92\,km for GenCast and 393\,km for IFS ENS. In Fig.~\ref{fig:forecast}c, one member each of \model,
GenCast and FourCastNet~3 keeps the storm's spiral of gales at day 5 and remains a sharp weather map, whereas the
deterministic forecasts of GraphCast\cite{lam2023}, FuXi\cite{chen2023} and Pangu-Weather\cite{bi2023} are smoother,
particularly at day 10. In Fig.~\ref{fig:forecast}b, the median RMSE improvement over the IFS ENS ensemble mean across 12 targets is 10.3\% for \model\ and 9.6\% for GenCast at day 3, 6.0\% for both at day 5 and 2.1\% and 2.3\% at day 10. The improvement of the deterministic models falls below theirs and, by day 10, below zero.

Power spectra show that \model's members keep the energy of ERA5 at short wavelengths. They match ERA5's spectra to
shorter wavelengths than the IFS ENS members do, for every field and lead time in Extended Data Fig.~\ref{ed:spectra}b,c. In the 2\,m temperature forecast over the Alps in Extended Data Fig.~\ref{ed:spectra}a, one
\model\ member retains the structure associated with mountains and coastlines.

\subsection{Skill across targets and lead times}
The skill in Fig.~\ref{fig:efficiency} is an average over a few variables and ten days. Here we compare \model\ with IFS ENS and GenCast for each target and lead time, and in the spread of the ensemble.
WeatherBench~2 scores geopotential (z), temperature (t), specific humidity (q) and the two wind components (u, v) at 500, 700 and 850\,hPa, and five surface variables: 2\,m temperature (t2m), mean sea-level pressure (msl), the two 10\,m wind components (u10, v10) and 10\,m wind speed (ws10). The CRPS of seven of these targets and of 24\,h precipitation is given in Fig.~\ref{fig:scorecard}a for IFS ENS, GenCast and \model. From day 3, the CRPS of \model\ and GenCast there agrees to within 2.5\% for every target, and \model's is the lower in 23 of the 32 cells.

\ifsubmission\else\ifsubmission
\begin{figure}[!ht]
\else
\begin{figure}[!t]
\fi
\centering
\includegraphics[width=\textwidth]{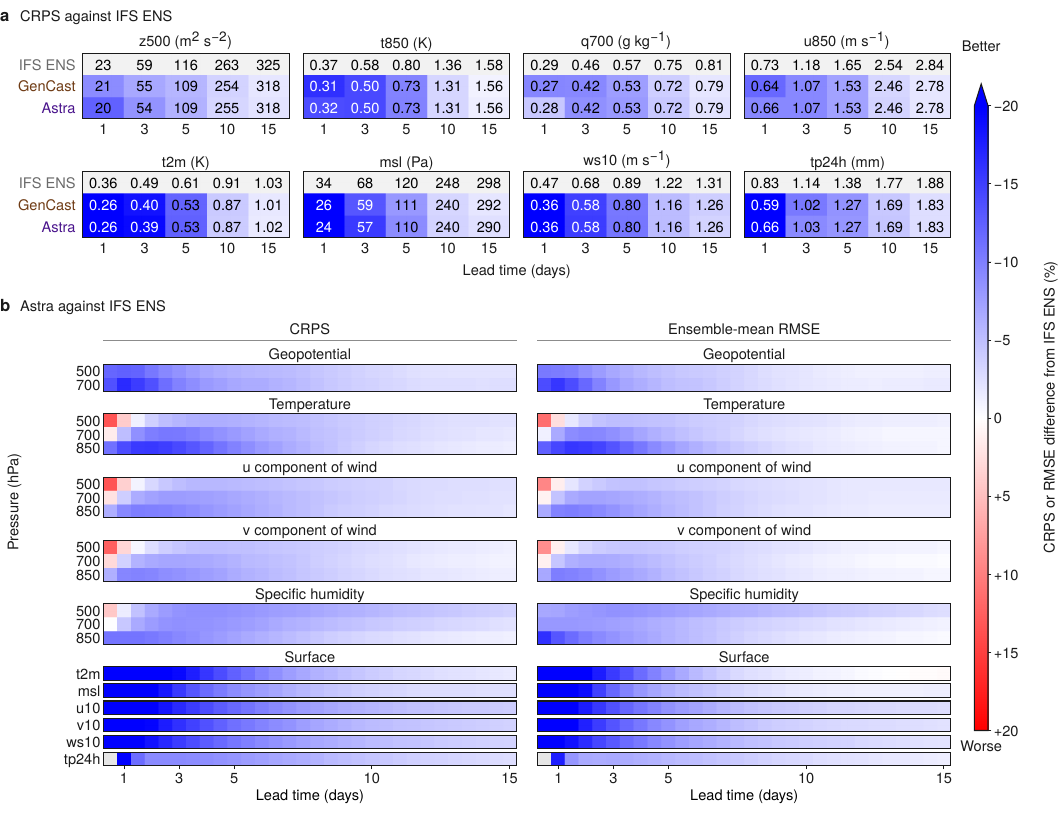}
\par\smallskip
\begin{minipage}{\textwidth}\footnotesize
\refstepcounter{figure}\label{fig:scorecard}%
\input{figures/fig_scorecard3/legend}
\end{minipage}
\end{figure}

\fi%

\model\ improves on IFS ENS in both the forecast distribution and its mean. Over the 30 lead times from 12\,h to 15~days, \model\ achieves a better CRPS than IFS ENS on 560 of the 570 combinations in Fig.~\ref{fig:scorecard}b and a better ensemble-mean RMSE on 558. Beyond the first 24\,h, \model's CRPS is better on every combination. When every model is verified against ERA5 instead, \model\ achieves better scores than IFS ENS on every combination, 850\,hPa geopotential included. This holds in our evaluation and when the forecasts are scored by WeatherBench~2, as Extended Data
Figs.~\ref{ed:scorecard_full}c,d and~\ref{ed:scorecard_1p5}c,d show.

\model\ and GenCast are statistically indistinguishable on 85\% of combinations in CRPS and on 87\% in ensemble-mean RMSE. At lead times of 3 days or more the significant differences in Extended Data Fig.~\ref{ed:scorecard_full}a,b all favour \model: 12 of 500 in CRPS and 13 in ensemble-mean RMSE. When the forecasts are scored by WeatherBench~2, \model's CRPS is lower than GenCast's on 513 of the 600 combinations and its ensemble-mean RMSE on 392; see e.g., Extended Data Fig.~\ref{ed:scorecard_1p5}a,b. By region, \model\ has a lower CRPS than GenCast in 873 of the 1,134 combinations of region, variable and lead time in Extended Data Fig.~\ref{ed:regions_1p5}a--f.

\model's ensemble spread grows to match its error. The spread--skill ratio is the ensemble spread divided by the ensemble-mean RMSE. Averaged over z500, t850, q700, u850, v850, t2m, msl and ws10
against ERA5, the ratio of \model\ rises from 0.90 at day 1 to 1.00 at day 15. At lead times of 9 days or more, it is closer to 1 than the ratios of GenCast and IFS ENS. At day 15, \model's ratio lies between 0.97 and 1.03 for each of z500, t850, q700, msl, t2m, ws10 and 24\,h precipitation in Extended Data Fig.~\ref{ed:curves}c. Extended Data Fig.~\ref{ed:curves}a,b shows the CRPS and RMSE skill of the same variables at every lead time.

\subsection{Tropical cyclones and precipitation}
\ifsubmission\else\ifsubmission
\begin{figure}[!ht]
\else
\begin{figure}[!tp]
\fi
\centering
\includegraphics[width=\textwidth,height=0.88\textheight,keepaspectratio]{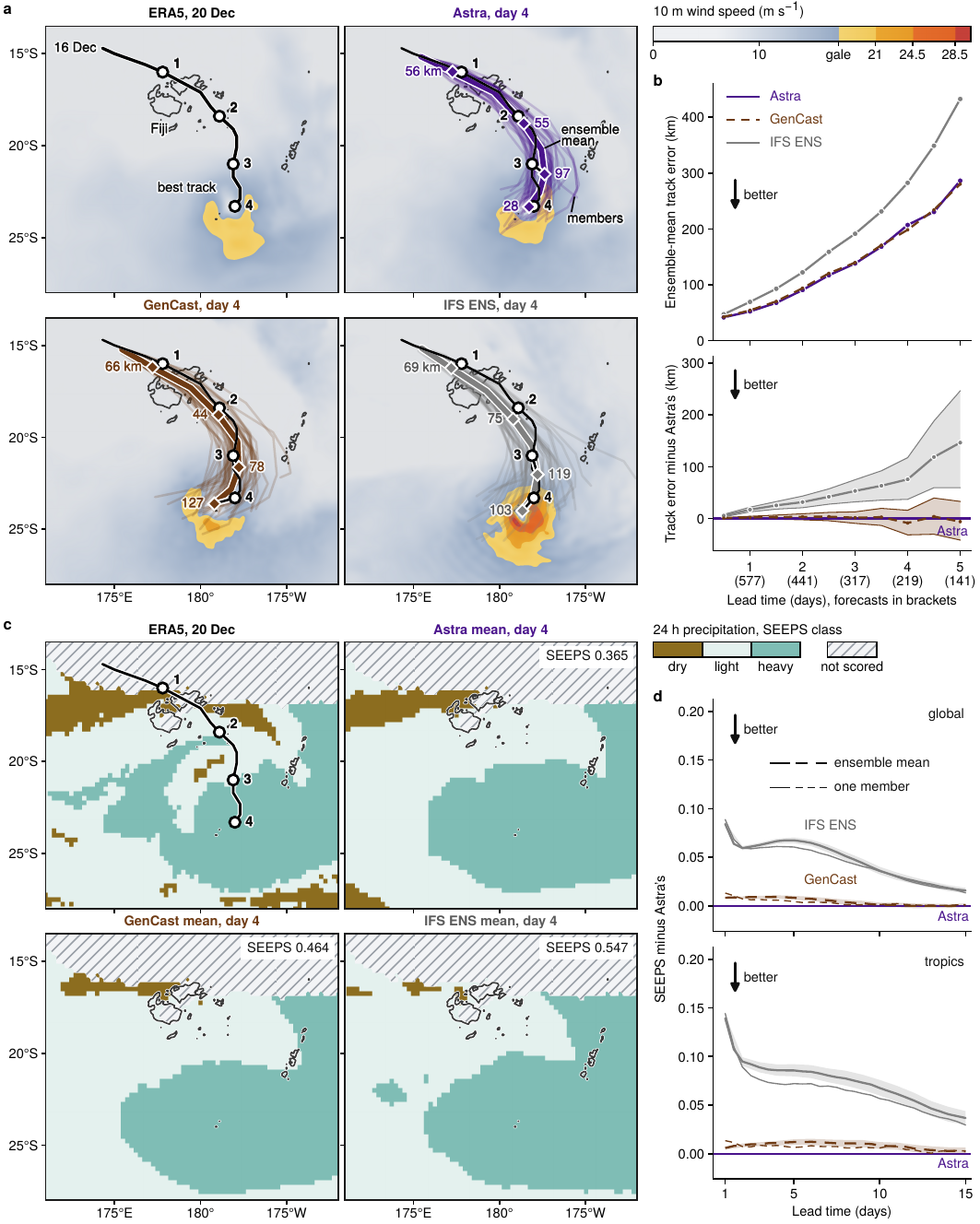}
\par\smallskip
\begin{minipage}{\textwidth}\footnotesize
\refstepcounter{figure}\label{fig:extreme}%
\textbf{Fig.~\thefigure\ \textbar\ \model's skill extends to tropical cyclone tracks and precipitation.}
\textbf{a},~Tropical Cyclone Yasa, forecast from 00\,UTC on 16 December 2020. Shading shows 10\,m wind speed in
ERA5 and in one member of each ensemble, in gold from gale force (17.5\,m\,s$^{-1}$). Thin lines are the tracks of the members. The black line is the IBTrACS
best track, with a numbered disc on each day. Diamonds mark the 50-member ensemble mean on each day; each is joined
to that day's disc and labelled with the distance in km.
\textbf{b},~Top, ensemble-mean track error (786 forecasts of 106 tropical cyclones in 2020). Bottom, the track
errors of GenCast and of IFS ENS minus that of \model; bands show 95\% bootstrap confidence intervals over cyclones.
\textbf{c},~Precipitation in the 24\,h to 00\,UTC on 20 December in the SEEPS\cite{rodwell2010} classes (dry, below
0.25\,mm), in ERA5 and in each system's 50-member mean, with the SEEPS over the mapped area.
\textbf{d},~SEEPS of 24\,h precipitation in 2020 (WeatherBench~2) of GenCast and of IFS ENS minus that of \model,
for the ensemble mean and for one member, over the globe and the tropics (20\textdegree\,S--20\textdegree\,N). Bands show 95\% block-bootstrap confidence intervals over initializations for the
ensemble mean.
\end{minipage}
\end{figure}

\fi%
\model\ tracks tropical cyclones as closely as GenCast and more closely than IFS ENS. For Cyclone Yasa in Fig.~\ref{fig:extreme}a, \model's ensemble-mean centre lies 28\,km from the best track at day 4, against 127\,km for GenCast and 103\,km for IFS ENS. Over the 106 tropical cyclones of 2020 in Fig.~\ref{fig:extreme}b, its ensemble-mean track error at lead times up to 5 days is comparable to that of GenCast
and significantly smaller than that of IFS ENS, by 146\,km at day 5.

\model\ forecasts precipitation better than IFS ENS. We score 24\,h precipitation with the stable equitable error in probability space
(SEEPS)\cite{rodwell2010}, which sorts rain into dry, light and heavy classes. For Cyclone Yasa in Fig.~\ref{fig:extreme}c, the SEEPS of \model's ensemble mean is the lowest, at 0.365, against 0.464 for GenCast and 0.547 for IFS ENS. In Fig.~\ref{fig:extreme}d, the ensemble mean of \model\ achieves a significantly better SEEPS than IFS ENS
at every lead time. It is also significantly better than that of GenCast up to day 9, and up to day 12 in the tropics. The CRPS of \model\ for 24\,h precipitation in Extended Data Fig.~\ref{ed:scorecard_full}c is also lower than that of IFS
ENS at every lead time.

\ifsubmission\else\ifsubmission
\begin{figure}[!ht]
\else
\begin{figure}[!tp]
\fi
\centering
\includegraphics[width=\textwidth]{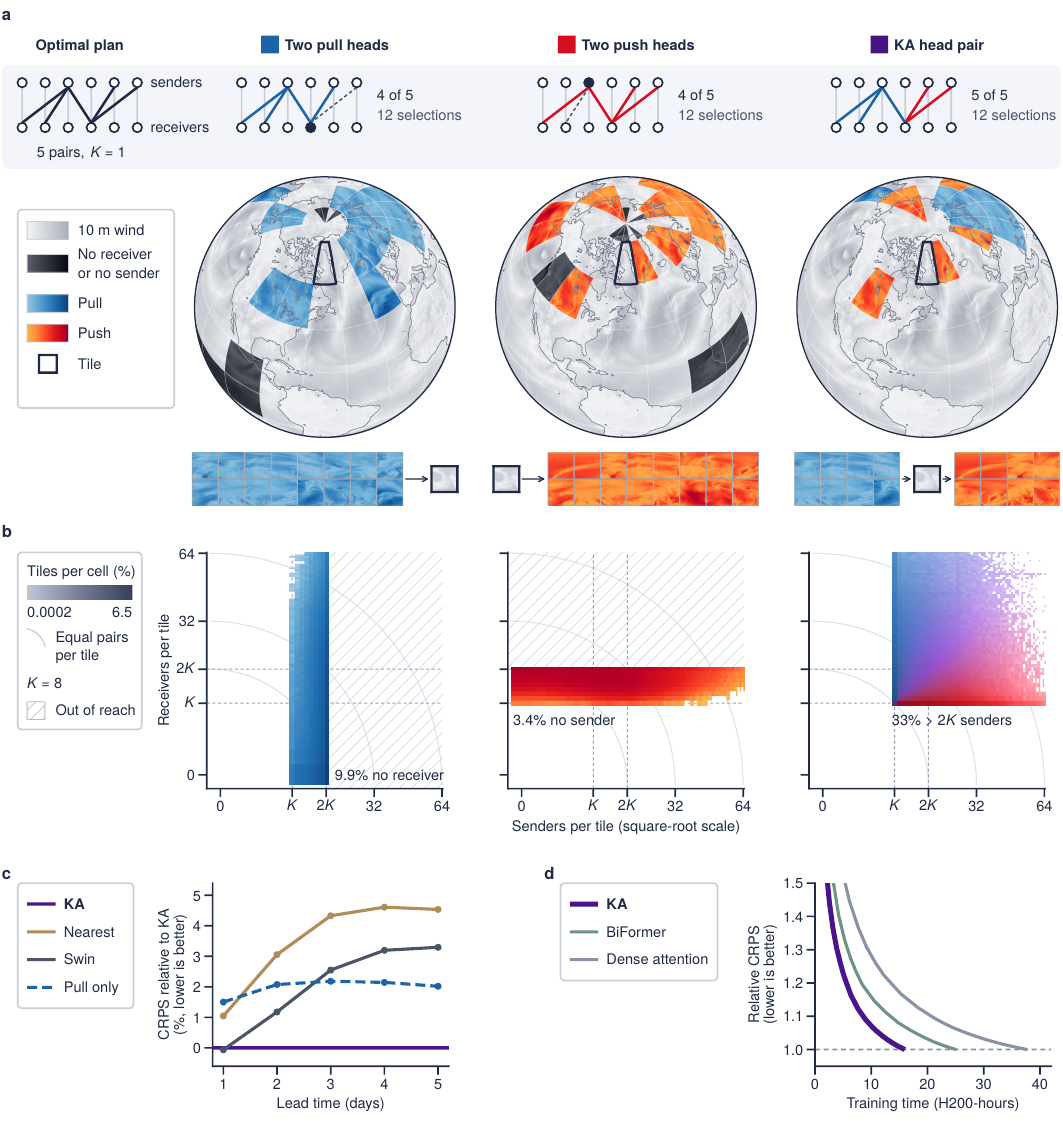}
\par\smallskip
\begin{minipage}{\textwidth}\footnotesize
\refstepcounter{figure}\label{fig:ka}%
\input{figures/fig_ka/legend}
\end{minipage}
\end{figure}

\fi%
\subsection{\hla}
\hla\ lets any two tiles on the globe exchange information at the attention cost of a local window per branch. Each tile selects both its senders and its receivers: in Fig.~\ref{fig:ka}a, each tile selects $K$ senders to receive from, in a pull branch, and $K$ receivers to send to, in a push branch. Half of the heads of a block pull and half push, and a head pair is one pull and one push head of the same index. With $K=8$, each branch processes as many tile pairs as a $3\times3$ window, the tile itself included, plus a small cost for scoring all $T^2$ pairs of the $T$ tiles. By the operation counts in section~S2 of the Supplementary Note, dense attention in its place would multiply the network's operation count 4.6-fold.

When both senders and receivers select $K$ tiles, KA can represent pairings that the same number of selections on one side cannot. With selection at the receiver alone, a tile can be selected by no receiver, and an optimal transport plan in which many tiles deliver to one cannot be kept, whatever scores are learned. In KA every tile sends to at least $K$ tiles, and by Supplementary Theorem~\ref{thm:ka_support} one head pair with suitable scores keeps an optimal plan for each of any $K$ linear transport problems. In the trained model, pull alone leaves 38\% of tiles at 1\textdegree\ and 28\% at 2\textdegree\ selected by no receiver in one head (Supplementary Note, section~S2). In Fig.~\ref{fig:ka}b, 33\% of the tiles of a head pair at 2\textdegree\ receive from more than $2K$ tiles, beyond the reach of two pull heads.

At a fixed operation count, KA forecasts better after 7,000 updates than three other rules for selecting tiles. In a quarter-width W-Net, we replaced KA's selection of tiles by the eight nearest tiles, the $3\times3$-tile windows of the Swin Transformer\cite{liu2021swin} or pull only, in which both branches select at the receiver. In Fig.~\ref{fig:ka}c, each of the three gives a significantly worse CRPS than KA over days 1--5: by 3.5\% with the nearest tiles, 2.0\% with the windows and 2.0\% with pull only. KA's lead over the nearest tiles and the windows grows with lead time, to 4.5\% and 3.3\% at day 5.

KA reaches the same CRPS 1.6 times faster than BiFormer's attention and 2.4 times faster than dense attention. We trained the quarter-width W-Net for longer with BiFormer's released attention module\cite{zhu2023biformer} in place of KA, or with dense attention in its place at 2\textdegree, with the settings given in Methods. In Fig.~\ref{fig:ka}d, KA reaches the CRPS at which its training plateaus in 16 H200-hours, BiFormer's attention in 25 and dense attention in 38. The three need about the same number of updates, and an update takes 2.3\,s with KA against 3.8\,s and 5.5\,s.

\subsection{Self-featurized CRPS}
\ifsubmission\else\ifsubmission
\begin{figure}[!ht]
\else
\begin{figure}[!tp]
\fi
\centering
\includegraphics[width=\textwidth]{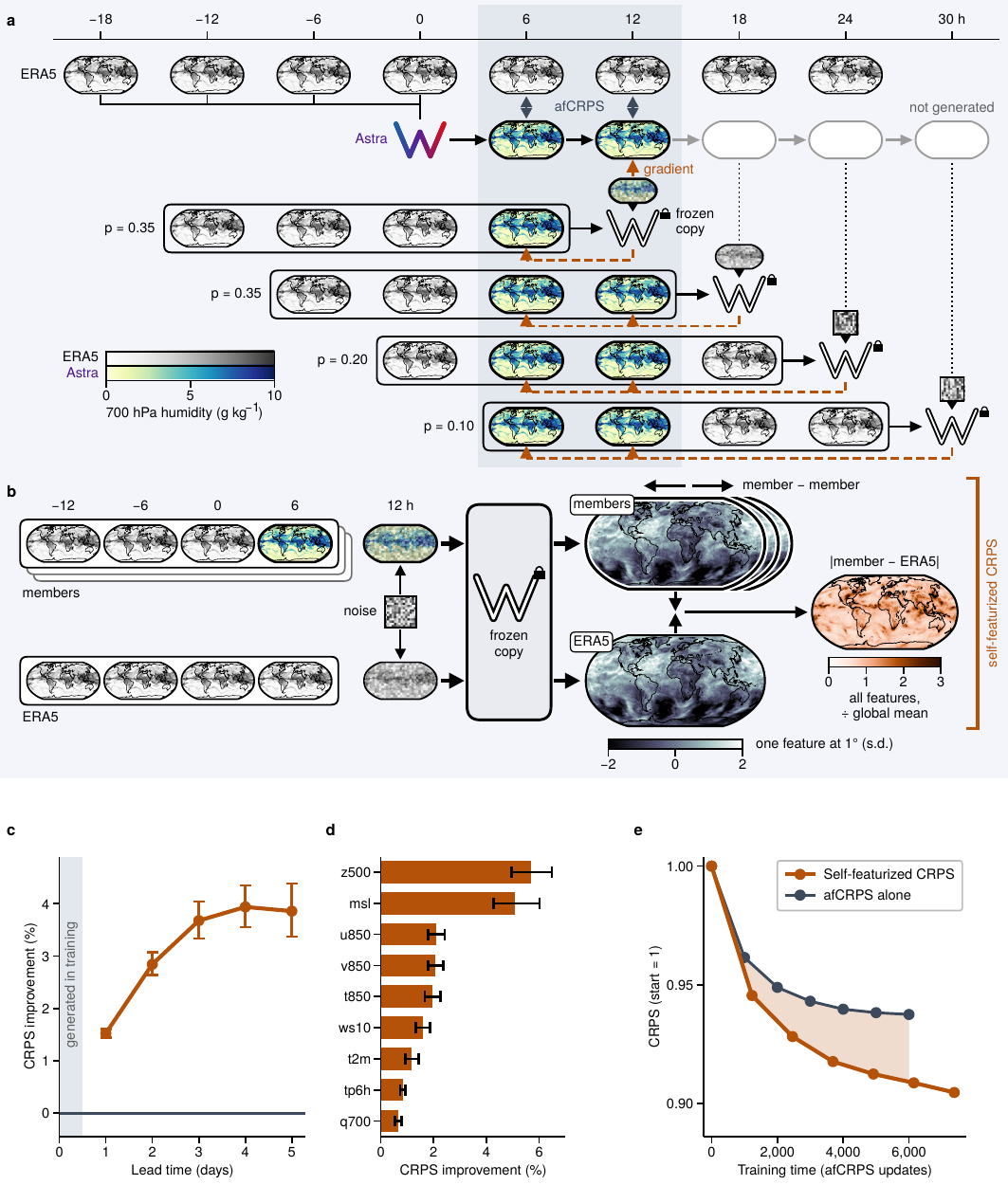}
\par\smallskip
\begin{minipage}{\textwidth}\footnotesize
\refstepcounter{figure}\label{fig:sf}%
\textbf{Fig.~\thefigure\ \textbar\ Self-featurized CRPS (SF-CRPS) improves forecasts one to five days ahead over almost-fair CRPS (afCRPS).}
\textbf{a},~In training, \model\ generates the states at 6 and 12\,h (grey shading), scored with afCRPS. A frozen copy processes the input window of one later step, with a lead time of 12, 18, 24 or 30\,h (boxed; drawn with probability $p$). It also takes a noised next state, the small map above each copy: the member's prediction at 12\,h, ERA5's state at 18\,h and noise alone at 24 and 30\,h. ERA5 fills the other states of each window.
\textbf{b},~For the first input window of \textbf{a}, the copy processes the members' predictions and ERA5 separately under the same noise. CRPS draws the members' features towards those of ERA5 and apart from one another. The map on the right shows the mean absolute difference between the two at 1\textdegree, divided by its global mean.
\textbf{c},~Improvement in CRPS from SF-CRPS over afCRPS alone, averaged over the 75 state variables
and 10\,m wind speed.
\textbf{d},~As \textbf{c}, for nine variables over 24--120\,h. tp6h is 6\,h precipitation, and the other abbreviations are as in Fig.~\ref{fig:scorecard}.
\textbf{e},~CRPS against training time, relative to its value at the start and averaged as in \textbf{c}. Training time is in afCRPS updates, and one SF-CRPS update counts as 1.23. Shading shows the gap between the two objectives.
\textbf{c}--\textbf{e},~Quarter-width W-Net; 16 members, 96 initializations of 2020. Error bars in \textbf{c} and
\textbf{d} show 95\% block-bootstrap confidence intervals over initializations.
\end{minipage}
\end{figure}

\fi%
SF-CRPS uses a generated rollout of only two steps and still improves forecasts days ahead. Models trained with CRPS usually learn later lead times from longer generated rollouts: up to eight in FourCastNet~3 and FGN and up to 12 in AIFS-CRPS\cite{bonev2025,alet2025,lang2024}. Each generated step adds a network evaluation and its backpropagation.

SF-CRPS extends the supervision of two generated steps to 30\,h, a lead time that a generated rollout reaches only after five steps. SF-CRPS scores each prediction twice: by the afCRPS\cite{lang2024} of its fields, and by the CRPS of the features that \model\ itself computes when it takes the prediction as input. These features are \model's own, so no separately trained feature network is needed. Each combines several variables and neighbouring locations, so its CRPS also reflects how they relate, which a score applied to each grid point and variable cannot. Because \model\ takes its four latest states as input, each predicted state remains an input for the next four steps.
In Fig.~\ref{fig:sf}a,b, a frozen copy of \model\ processes the predictions in the inputs of these subsequent steps, with ERA5 supplying the remaining states. The two generated states, at 6\,h and 12\,h, are thus supervised in the inputs of steps up to 30\,h ahead.

SF-CRPS improves CRPS at every lead time (Fig.~\ref{fig:sf}c) and for every variable (Fig.~\ref{fig:sf}d). We trained the quarter-width W-Net from the MeanFlow weights of the network with KA once with SF-CRPS and once with afCRPS. Averaged over the 75 state variables and 10\,m wind speed, the gain of SF-CRPS grows from 1.5\% at 24\,h to 2.8\% at 48\,h, 3.7\% at 72\,h and 3.9\% at 96\,h and 120\,h, well beyond the 30\,h that the loss covers. Its 95\% confidence interval lies above zero at every lead time. Averaged over these lead times, the gain is 5.7\% for geopotential at 500\,hPa and 5.1\% for mean sea-level pressure. It is 2.0--2.1\% for temperature and the winds at 850\,hPa and 0.7--1.6\% for 10\,m wind speed, 2\,m temperature, 6\,h precipitation and specific humidity at 700\,hPa. The largest gains are in large-scale fields, in line with the copy's features, which come from the network's four coarsest levels.

SF-CRPS also keeps improving where afCRPS stops. In Fig.~\ref{fig:sf}e, training with afCRPS plateaus: over its last 1,000 updates, CRPS falls by 0.08\% of its starting value, against 0.41\% with SF-CRPS. An SF-CRPS update takes 1.23 times as long as an afCRPS update, and at equal training time the gain is still 3.0\%. SF-CRPS is also better in each of three seeds of a shorter training run, by 3.7\% on average (95\% confidence interval, 2.8--4.6\%). Because the frozen copy is used only in training, these gains leave the forecasting network unchanged and add no cost at inference.

\section{Discussion}

Our results show that accurate 0.25\textdegree{} ensemble forecasting
can be achieved with substantially lower computational cost.
On the 2020 benchmark, \model\ achieves aggregate forecast skill
comparable to GenCast using only 9.8\% of its
training cost, while retaining fine-scale spatial structure.
Training uses one eight-GPU node, and a 50-member, 15-day forecast
takes 34 minutes on a single consumer GPU.

The two methodological contributions address distinct sources of computational cost. \hla\ reduces the cost of global, state-dependent information exchange while allowing each tile to select both its senders and receivers. At matched operation counts, it improves forecast skill over position-based and receiver-only selection; in longer runs, it reaches the same CRPS as BiFormer's attention or dense attention in less training time.
Its selected pairs can support optimal transport plans that one-sided selection cannot guarantee. SF-CRPS uses features from a frozen copy of the forecasting model to evaluate predictions in subsequent forecast inputs, improving forecasts one to five days ahead over afCRPS from only two generated 6\,h steps. This improvement adds no inference cost. Generating each forecast step in one network evaluation, as FGN and FourCastNet~3 also do\cite{alet2025,bonev2025}, further reduces the cost of producing large ensembles.

The next steps for \model\ are forecasts initialized from operational analyses and extensions to grids finer than 0.25\textdegree. More broadly, global information exchange and autoregressive prediction are common to models across science and engineering. Two-sided selection and supervision through a model's own features could therefore offer complementary routes to more efficient learning in these settings. For weather forecasting, the present results lower the computational barrier for more research groups to develop, evaluate and refine 0.25\textdegree{} ensemble models.

\section{Methods}
\subsection{Forecast task and data}
\model\ maps a draw of noise to the 6\,h residual, the change of the state over the next 6\,h, in one network evaluation. ERA5\cite{hersbach2020}, as provided by WeatherBench~2\cite{rasp2024}, supplies training data for 1979--2019 and evaluation data for 2020 on the 0.25\textdegree\ grid of $721\times1{,}440$ points. The 75-variable state comprises five atmospheric fields on 13 pressure levels, four surface fields and six auxiliary fields, which Extended Data Table~\ref{edt:vars}b lists. \model\ takes the four latest states 6\,h apart together with fields that it does not forecast. These are 36 static fields such as orography and the land--sea mask, and 13 forcing fields such as the insolation and the time of day and year. Details are in Supplementary Note, section S5.

Three forcing fields describe the surface: the sea-surface temperature anomaly, sea-ice cover and top-layer soil water. During a forecast they keep their 00\,UTC values of the initialization day. A further input gives the age of these values and grows with lead time. The normalization statistics and the masks were computed from ERA5 data of 1979--2019.

The W-Net has five resolution levels and applies two U-Nets\cite{ronneberger2015,newell2016} in sequence: encoder~1, decoder~1, encoder~2 and decoder~2, as Extended Data Fig.~\ref{ed:arch}a shows. The multiscale features of decoder~1 pass to the second U-Net. SF-CRPS also scores these features, as described below. The network takes the two MeanFlow times $(r,\tau)$, introduced below, as inputs. They set the scale and shift of its Derf layers\cite{chen2025derf}, the score gates of KA and a rotation of the attention queries. Derf layers are point-wise functions that take the place of normalization layers. The architecture is detailed in Extended Data Table~\ref{edt:cost}a and Supplementary Note, section S4.

\subsection{\hla\ with selection on both sides}
\hla\ divides the tokens of a level into $T$ tiles and exchanges tokens among them through a pull branch and a push branch. Each tile $i$ has two descriptors pooled from its tokens: a query descriptor $\hat{\mathbf q}_i$, used when the tile is a receiver, and a key descriptor $\hat{\mathbf k}_i$, used when it is a sender. In \model, each tile selects $K=8$ other tiles in every head and counts itself as a ninth. In a pull head the tile receives from these nine as its senders, and in a push head it sends to these nine as its receivers. Each branch has its own projections and routing scores $s_{ij}$. In both branches, the receiver supplies the queries and the sender supplies the keys and values:
\begin{equation}
s_{ij} = \frac{\hat{\mathbf{q}}_i^{\top}\hat{\mathbf{k}}_j}{\sqrt{d_k}},
\qquad
\mathbf{Y}_i = \operatorname{softmax}\!\Bigl(
  \frac{\mathbf{Q}_i \mathbf{K}_{\Pi(i)}^{\top}}{\sqrt{d_k}} \Bigr)
  \bigl( \mathbf{G}_i \odot \mathbf{V}_{\Pi(i)} \bigr),
\label{eq:transport}
\end{equation}
where $\mathbf Q_i$ holds the queries of the tokens of tile $i$, $\mathbf K_{\Pi(i)}$ and $\mathbf V_{\Pi(i)}$ hold the keys and values of the tokens of the tiles in $\Pi(i)$, and $d_k$ is the head dimension. The scores thus compare tiles, whereas the softmax acts on tokens. The set $\Pi(i)$ holds the tiles whose tokens reach receiver $i$, and $\mathbf G_i$ gates the values of sender $j$ by a learned function $g$ of the score $s_{ij}$. The two branches differ only in $\Pi(i)$:
\begin{equation}
\begin{aligned}
\text{pull:}\quad \Pi(i)&=\mathcal N(i)=\{i\}\cup\operatorname{top}_K\{s_{ij}\}_{j\neq i},\\
\text{push:}\quad \Pi(i)&=\mathcal S(i)=\{\,j:i\in\mathcal R(j)\,\},\qquad \mathcal R(j)=\{j\}\cup\operatorname{top}_K\{s_{i'j}\}_{i'\neq j},
\end{aligned}
\label{eq:graph}
\end{equation}
where $\operatorname{top}_K$ takes the $K$ tiles with the highest scores.

\textbf{Guarantees.} Selecting on both sides with separate scores guarantees three properties of what KA can represent. We describe what it can represent by linear transport problems. Each gives every receiver and every sender a positive mass and every pair a cost, and an optimal plan moves the masses at the least total cost\cite{kantorovich1942,peyre2019ot}. Let a head pair denote one pull head and one push head of the same index, which score and select independently; it supports a plan if the plan moves mass only along the pairs that either head selects or between a tile and itself.

First, a head pair supports any $K$ problems, and that is the most two heads can do. An optimal plan can use $2T-1$ pairs, about two for each tile. A head pair selects $2K$ pairs for each tile, $2TK$ in all, and with suitable scores these contain an optimal plan for each of any $K$ problems (Supplementary Theorem~\ref{thm:ka_support}). One more problem can push the pairs needed beyond $2TK$, so some $K+1$ problems cannot be supported by any two heads, however they select (Supplementary Proposition~\ref{prop:ka_budget}). The $h$ head pairs of a block thus support exactly $hK$ problems: 24 at 1\textdegree\ and 48 at 2\textdegree, as Extended Data Table~\ref{edt:ka_hold} lists.

Second, this needs selection on both sides. Heads that all select at the receiver, or all at the sender, cannot support an optimal plan for some single problem, whatever their scores (Supplementary Theorem~\ref{thm:ka_support}). In that problem one tile receives from every other tile and sends to every other tile, and selection on one side gives the tile only $K$ of them in a head.

Third, pull and push must not share their scores. If both heads selected with one score matrix, a head pair could support only $\lfloor(K+1)/2\rfloor$ problems, four instead of eight for $K=8$ (Supplementary Proposition~\ref{prop:ka_shared}). KA therefore splits the heads of multi-head attention: half of them pull and half push, each with its own scores. Section~S1 of the Supplementary Note characterizes exactly which sets of pairs a head pair can select with separate scores and with shared scores.

Two synthetic tasks show both needs in practice. We compare KA with pull only, a variant that keeps KA's budget but has both branches select at the receiver, on the synthetic transport task of Supplementary Note, section~S3. With 49 or more sources per receiver, the error of pull only is 58\% of that of an all-zero output, against 1.1\% for KA. In the learned routing task of Extended Data Fig.~\ref{ed:routing_toy}b, pull only likewise loses delivered pairs as the skew grows, down to half. In the same routing task, a head pair with shared scores recalls either the requested pairs or the delivered pairs in full and no more than 10\% of the other.

The implementation of KA has these further details. The tile descriptors have their mean over all tiles subtracted, which removes the parts of the score that are common to all receivers or to all senders. A push head computes its softmax in parts: each sender forms one message for each of its selected receivers, and merging the messages at a receiver gives exactly equation~\eqref{eq:transport}\cite{milakov2018,dao2022flash}. Either branch therefore processes $T(K+1)$ pairs of tiles per head. Tiles that a receiver does not select contribute through a pooled correction, the weighted mean of their values minus that of the selected tiles. The outputs of all heads are concatenated and projected, and alternate blocks shift the tiles by half a tile\cite{liu2021swin}. The full specification is in Supplementary Note, section S1, and the tile sizes and the numbers of tiles and heads are in Extended Data Tables~\ref{edt:cost}a and~\ref{edt:ka_hold}.

\subsection{Forecast map and two-stage training}
Training has two stages, MeanFlow and then SF-CRPS. The first stage uses MeanFlow\cite{geng2025} to learn the average-velocity network $\mathbf u_\theta(\mathbf z_\tau,r,\tau)$. Here $\mathbf z_\tau$ interpolates between the standardized 6\,h residual at $\tau=0$ and the noise $\boldsymbol\varepsilon$ at $\tau=1$, and the network predicts the average velocity of this flow between the times $r$ and $\tau$. One evaluation over the whole flow, $(r,\tau)=(0,1)$, gives the forecast map
\begin{equation}
\mathbf{x}_{t_0+6\,\mathrm{h}} = \mathbf{x}_{t_0} +
\boldsymbol{\sigma}_\Delta\bigl(\boldsymbol{\varepsilon} -
\mathbf{u}_\theta(\boldsymbol{\varepsilon}, 0, 1)\bigr),
\label{eq:forecast}
\end{equation}
where $\boldsymbol\sigma_\Delta$ is the per-variable standard deviation of 6\,h residuals. The sampling of the flow times and the weighting of the MeanFlow loss are given in Supplementary Note, section S6.

The prior of the noise $\boldsymbol\varepsilon$ is fitted to the spatial spectra, cross-variable correlations and wind balance of ERA5 residuals of 1998--2019, as Extended Data Fig.~\ref{ed:prior} shows. Both stages draw noise from this prior. The five slowly varying auxiliary fields receive no noise.

The first stage takes 40,000 updates, and the second starts from its weights and takes 15,000, with schedule-free\cite{defazio2024} Muown\cite{lion2026muown}. Computation uses bfloat16, and the losses are evaluated in 32-bit floating point. Forecasts for evaluation use the running average of the weights at the end of training. The remaining settings are in Extended Data Table~\ref{edt:hyp}c and Supplementary Note, section S6. The conversion of training costs to V100-days and to costs per grid point is described in Supplementary Note, section S7.

\subsection{Self-featurized CRPS in the inputs of subsequent steps}
The second training stage, SF-CRPS, applies the forecast map of equation~\eqref{eq:forecast} for $G=2$ successive steps of 6\,h, with $M$ members. We write $\hat x_k$ for the states generated by member $k$ and $y$ for the ERA5 states at the same times. Step $s$ forecasts the lead time of $6s$\,h from the $H=4$ states that precede it and from the noised residual $\mathbf z_\tau$ of the MeanFlow. The training objective is a weighted sum of a field term and a feature term.

\textbf{Field term.} The outputs at 6\,h and 12\,h are scored with almost-fair CRPS\cite{lang2024},
\begin{equation}
\mathrm{afCRPS}_\alpha\bigl(\{x_k\};y\bigr) = \frac{1}{M}\sum_{k}\lvert x_k - y\rvert
 - \frac{1 - (1-\alpha)/M}{2M(M-1)}\sum_{k\neq k'}\lvert x_k - x_{k'}\rvert ,
\qquad \alpha = 0.95,
\label{eq:afcrps}
\end{equation}
where the $x_k$ are the values of the $M$ members and $y$ is the target. At $\alpha=1$ this is fair CRPS\cite{ferro2014}. The field term applies $\mathrm{afCRPS}_{0.95}$ to each of five scale bands of the fields, obtained with a Laplacian pyramid from 0.25\textdegree\ to 4\textdegree, as specified in Supplementary Note, section S6.

\textbf{Feature term.} Self-perceptual losses can use the frozen features of the generator itself\cite{lin2025}. Here a frozen copy of the model scores the generated states as inputs of subsequent steps. We write $\varphi_{s,l}(x)$ for the feature of the frozen copy at level $l$ in step $s$, when the states $x$ take the places of the generated states among the inputs of this step and ERA5 supplies the rest. The feature at level $l$ is the output of decoder~1, for the levels L1--L4. The copy uses the same noise and the same flow times for the members and for ERA5. The feature term is
\begin{equation}
\mathcal L_{\mathrm{feat}} = \mathbb E_{s}\sum_{l} w_l\,
\mathrm{afCRPS}_1\bigl(\{\varphi_{s,l}(\hat x_k)\};\,\varphi_{s,l}(y)\bigr),
\qquad s\in\{2,\ldots,G+H-1\},
\label{eq:sfcrps}
\end{equation}
with level weights $w_l$, evaluated over the channels and grid points of each feature map with area weighting. Each update draws one step $s$. The states generated at 6\,h and 12\,h are thus scored at the steps with lead times of 12, 18, 24 and 30\,h, and the steps beyond 12\,h are never generated in training.

Gradients pass through the frozen copy and through both generated steps, while the parameters of the copy stay fixed. The copy is refreshed from the running average of the model weights. Each level of the feature term is divided by the mean absolute error of its features, treated as a constant, which keeps the gradients of the levels comparable. The weights of the two terms, the probabilities of the steps, the level weights $w_l$ and the refresh interval are listed in Extended Data Table~\ref{edt:hyp}c.

\subsection{Forecast evaluation and uncertainty}
\textbf{Forecasts and ground truth.} Following the evaluation protocol of WeatherBench~2\cite{rasp2024}, we evaluate 50-member, 15-day forecasts from all 732 initializations at 00 and 12\,UTC in 2020. Members share the four ERA5 initial states and draw fresh noise at each step. Each member is then rolled out autoregressively from its own outputs. For 1.5\textdegree\ scores, all 50 members are conservatively regridded before evaluation\cite{jones1999}. We use ERA5 as ground truth, with one exception. In Figs.~\ref{fig:efficiency}a,b, \ref{fig:forecast}b and~\ref{fig:scorecard}, IFS ENS is verified against its own analysis, following GenCast\cite{price2025}, except for 24\,h precipitation, which is verified against ERA5 for every system.

\textbf{Scores.} All scores are area-weighted and averaged over all initializations, and square roots and ratios are taken after the averaging. They are computed at 30 lead times from 12\,h to 15 days. CRPS is the fair estimator, equation~\eqref{eq:afcrps} with $\alpha=1$. Ensemble-mean RMSE uses $[\overline{(\bar x-y)^2}-\overline{\hat\sigma^2}/M]^{1/2}$, where $\bar x$ is the ensemble mean, the overline is this average and $\hat\sigma^2$ is the ensemble variance with denominator $M-1$. These estimators account for the unequal ensemble sizes: 50 members for \model\ and IFS ENS, 56 for GenCast. The spread--skill ratio divides $(\overline{\hat\sigma^2})^{1/2}$ by this RMSE. A value of 1 means that the spread matches the ensemble-mean error\cite{fortin2014}. The skill of a score $S$ relative to IFS ENS is $100\,(1-S/S_{\textrm{IFS ENS}})$, averaged over the variables and lead times stated with each figure. In Fig.~\ref{fig:forecast}b and for the deterministic models in Fig.~\ref{fig:efficiency}b, the ensemble-mean RMSE of IFS ENS is taken without the $\overline{\hat\sigma^2}/M$ term. Further details of the scores are given in Supplementary Note, section S8.

\textbf{Uncertainty.} We compute confidence intervals with a circular block bootstrap\cite{politis1992} of the time-ordered initializations, which handles their temporal dependence. Each of 10,000 resamples draws blocks of 14 initializations (7 days), and the intervals are two-sided 95\% percentile intervals. Two systems differ significantly when the interval of their difference excludes zero. For a ratio of scores, the scores are averaged in each resample before the ratio is taken. At 0.25\textdegree, where the scores of the other models are their published averages, only the scores of \model\ are resampled and the intervals are widened by a calibrated factor, as section~S9 of the Supplementary Note describes. These intervals describe variation across forecast initializations.

\textbf{Tropical cyclones.} For the storm of Fig.~\ref{fig:forecast}, we locate the centre in each member as the minimum of mean sea-level pressure within 8\textdegree\ of the centre in ERA5 at the same lead time, in 50 members of each ensemble. The position error is that of the ensemble-mean centre. For Fig.~\ref{fig:extreme}a,b, we track tropical cyclones with the same tracker in the three 50-member ensembles, on the 1.5\textdegree\ grid of the WeatherBench~2 evaluation. We treat the IBTrACS best tracks\cite{knapp2010} as ground truth. A forecast is counted for each cyclone that IBTrACS records at the initialization time with winds of at least 34\,kt and that is not extratropical. This gives 786 forecasts of 106 cyclones. Each member is tracked from the best-track position at the initialization time, every 12\,h. At each time the tracker extrapolates the position from the motion of the member and takes the nearest local minimum of mean sea-level pressure within 5\textdegree, refined by a quadratic fit to the surrounding grid points. Tracking continues while a 200\,Pa contour within 5.5\textdegree\ encloses the minimum\cite{ullrich2021}. The ensemble-mean position is defined when at least half of the members still track the cyclone, as in GenCast\cite{price2025}. We average the position error over the forecasts for which all three systems provide an ensemble-mean position. The 95\% confidence intervals come from a bootstrap that resamples cyclones 10,000 times.

\textbf{Precipitation.} Precipitation forecasts are scored with SEEPS\cite{rodwell2010} for 24\,h accumulations, as in WeatherBench~2. In Fig.~\ref{fig:extreme}c, SEEPS is computed over the mapped area. In Fig.~\ref{fig:extreme}d, it is the SEEPS of the WeatherBench~2 evaluation, for the ensemble mean and for the first member, over the globe and over the tropics (20\textdegree\,S--20\textdegree\,N). Extended Data Fig.~\ref{ed:scorecard_1p5}e lists the global values for the ensemble mean. The SEEPS comparisons of Fig.~\ref{fig:extreme}d are paired tests at 1.5\textdegree, with the SEEPS of each system computed per initialization.

\textbf{Spectra.} We say that a spectrum agrees with ERA5 down to a wavenumber if its power remains within a factor $\sqrt2$ of that of ERA5 at every degree from 1 to that wavenumber. We report the largest such wavenumber, as detailed in Supplementary Note, section S9.

\subsection{Ablations at a quarter of the width}
The ablations of Fig.~\ref{fig:ka}c,d and Fig.~\ref{fig:sf}c--e use the quarter-width W-Net, which has a quarter of the width of \model's network and the same tiles and $K$. Each model is trained on one H200 GPU on ERA5 of 2018--2019. The models of Fig.~\ref{fig:ka}c,d are trained with MeanFlow from random weights.

\textbf{Selection rules (Fig.~\ref{fig:ka}c).} The three selection rules replace only the choice of tiles. With the nearest tiles, each tile attends to its eight nearest tiles; with the windows of the Swin Transformer\cite{liu2021swin}, the tiles are grouped into windows of $3\times3$ tiles, shifted by half a window in alternate blocks, and each tile attends to the tiles of its window; with pull only, the push heads select at the receiver. Each rule attends over nine tiles per tile, the tile itself included, and its operation count is within 0.3\% of that of KA. KA and each of the three selection rules are trained from three seeds for 7,000 updates. CRPS is scored on 64 initializations of 2020 at 24--120\,h with 16 members. The difference from KA is 100 times the mean of the logarithm of the CRPS ratio, taken over z500, t850, q700, msl, u850, v850, t2m and ws10 and over the five lead times. Its 95\% confidence interval combines the variance between seeds, pooled over all comparisons of these runs, with the variance of a bootstrap over seven-day blocks of initializations.

\textbf{Attention modules (Fig.~\ref{fig:ka}d).} In the BiFormer variant, we replace KA at 1\textdegree\ and 2\textdegree\ with the bi-level routing attention module of the released BiFormer models\cite{zhu2023biformer}, with other layers kept the same. Each tile attends to the tokens of the 9 tiles with the highest affinity between tile-mean queries and keys. A dense block at 1\textdegree\ with KA's heads would take 109 times the operations of a KA block. The dense-attention variant therefore replaces KA only at 2\textdegree, with dense attention of 12 heads of width 128, and its forward pass takes 5.0 times the FLOPs of the network with KA. KA, BiFormer's attention and dense attention each train with one seed for 30,000 updates. Checkpoints are scored by the all-variable fair CRPS on 16 initializations of 2020 (24 and 48\,h, 16 members). Training times are update counts times the steady H200 time per update: 2.30, 3.80 and 5.50\,s for the KA, BiFormer and dense-attention variants. Training is taken as converged at the checkpoint after which two successive 1,000-update gains are at most 0.5\%: 25,000, 23,000 and 26,000 updates. The CRPS at these checkpoints agrees to within 0.6\%. Figure~\ref{fig:ka}d plots CRPS against training time, and for each variant we report the time at which its curve reaches the CRPS of KA at 25,000 updates.

\textbf{Training objectives (Fig.~\ref{fig:sf}c--e).} The afCRPS variant uses the field term alone. The SF-CRPS variant adds the feature term, and its frozen copy is the model at the start of the stage. The second stage starts from the MeanFlow weights of KA after 25,000 updates and runs for 6,000 updates, with one initialization and four members per update. Forecasts are scored on 96 initializations of 2020 at 24--120\,h with 16 members. For each of the 75 state variables and for 10\,m wind speed, the CRPS is averaged over the initializations, and the improvement is $1-\mathrm{CRPS}_{\text{SF-CRPS}}/\mathrm{CRPS}_{\text{afCRPS}}$. The improvement at a lead time is the mean of these 76 improvements. For Fig.~\ref{fig:sf}e, the CRPS at 24--120\,h of each checkpoint, saved every 1,000 updates, is divided by that of the MeanFlow weights and averaged over the same 76 variables. Training time counts one update with SF-CRPS as 1.23 updates with afCRPS. The value at equal training time is interpolated in the logarithm of CRPS between the checkpoints at 4,000 and 5,000 updates. The 95\% confidence intervals are percentiles of 4,000 resamples of 52 seven-day blocks. The comparison is repeated for three seeds with a shorter training run, and scored as for Fig.~\ref{fig:ka}c.

\subsection{Use of large language models}
Claude (Anthropic) was used to assist with preparing the figures, with checking the text and the proofs, and with coding. The authors reviewed all content and take responsibility for it.

\section{Data availability}
ERA5\cite{hersbach2020} is available from the Copernicus Climate Change Service. The WeatherBench~2 public
bucket\cite{rasp2024} provides ERA5 on its grids, the forecasts of GenCast, IFS ENS, GraphCast, FuXi and
Pangu-Weather, and result files with their scores and those of IFS HRES, HEAL-ViT and Baguan. IBTrACS best
tracks\cite{knapp2010} are available from NOAA
(\url{https://www.ncei.noaa.gov/products/international-best-track-archive}). The FourCastNet~3 member of
Fig.~\ref{fig:forecast}c was generated with its released weights (\url{https://huggingface.co/nvidia/fourcastnet3})
from an ERA5 initial state.
The 50-member forecasts for 2020, conservatively regridded from the native
0.25\textdegree\ outputs to the standard 1.5\textdegree\ WeatherBench~2 grid, are available at
\url{https://huggingface.co/datasets/JunuChoi/Astra-WB2-2020-240x121}
under the CC~BY~4.0 licence.

\section{Code availability}
Code for WeatherBench~2 scoring, scorecards and significance tests, with \model's per-initialization
0.25\textdegree\ scores, is available at \url{https://github.com/Junu-Choi/astra-wx}.
Code for the model, training and forecasting will be released in the same repository.
The repository uses the BSD-3-Clause-Clear licence, with Apache~2.0 applying to the \texttt{eval025} evaluation code.

\newcounter{mainrefs}\ifcsname c@NAT@ctr\endcsname\setcounter{mainrefs}{\value{NAT@ctr}}\else\setcounter{mainrefs}{\value{@bibitem}}\fi

\section{Acknowledgements}
We thank Stephan Rasp and the maintainers of WeatherBench~2 for evaluating our released forecasts and adding \model\ to the
benchmark.

\section{Funding}
The work of Y. Hong was supported by the ASTRA Project through the National Research Foundation (NRF) funded by the
Ministry of Science and ICT (No. RS-2024-00440063).

\section{Author contributions}
Y.H. and J.C. conceived and designed the study. J.C. developed the model, performed the experiments and analysed the
data. J.C. wrote the manuscript, and Y.H. revised it. Y.H. supervised the research and acquired funding. Both authors
discussed the results and approved the final manuscript.

\section{Competing interests}
The authors declare no competing interests.

\ifsubmission
\section{Figures and figure legends}
\ifsubmission\fi
\ifsubmission\fi
\ifsubmission\fi
\ifsubmission\fi
\ifsubmission\fi
\ifsubmission\fi
\clearpage
\fi

\clearpage
\section{Extended Data}
\par\noindent\begin{minipage}{\textwidth}
\edfig{The network: a five-level W-Net of 3.0 billion parameters, with \hla\ at 1\textdegree\ and 2\textdegree.}\label{ed:arch}
\noindent\makebox[\textwidth][c]{\includegraphics[width=180mm]{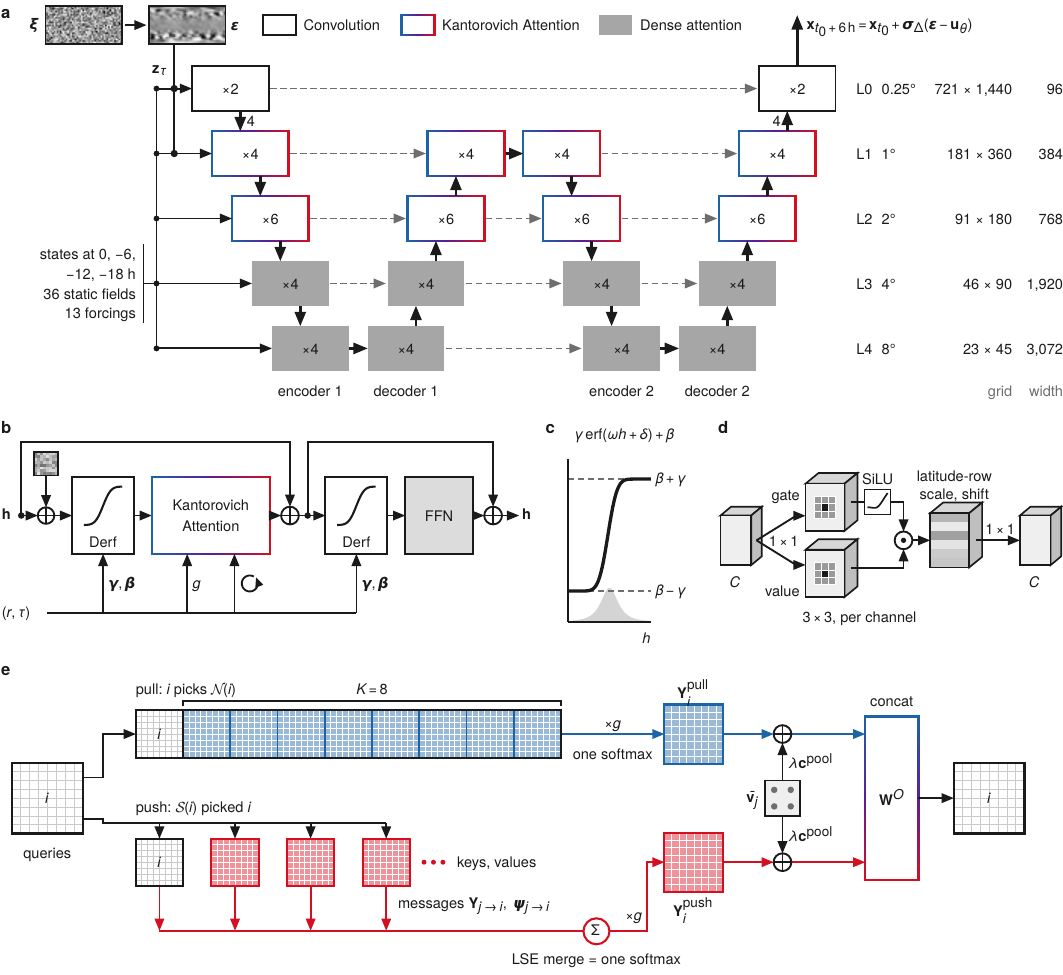}}\par\vspace{1.5mm}
\par\noindent{\footnotesize
\textbf{a}, The W-Net of Fig.~\ref{fig:forecast}d. Boxes give the number of blocks, vertical arrows resample by 2 or as marked, and dashed lines are skip connections. $\mathbf{z}_\tau$ is the noised 6\,h residual, which is the prior noise $\boldsymbol{\varepsilon}$ at inference, coloured from white noise $\boldsymbol{\xi}$.
\textbf{b}, The block at L1--L2, with \hla\ as in \textbf{e}. The grey tile is the learned tile positions, and the flow times $(r,\tau)$ set Derf's $\boldsymbol{\gamma}$ and $\boldsymbol{\beta}$, the score gates $g$ and the query rotation (circular arrow). L3--L4 use dense attention and L0 a $3\times3$ convolutional FFN.
\textbf{c}, Derf maps a heavy-tailed input $h$ (grey) to values between $\beta-\gamma$ and $\beta+\gamma$.
\textbf{d}, The SwiGLU FFN of L1--L4, with the latitude-row scale and shift at L1--L2 only; $C$ is the level's width.
\textbf{e}, One head pair for receiver tile $i$ (white), whose queries are the black arrows: the pull head (blue) attends over $\mathcal N(i)$ and the push head (red) over $\mathcal S(i)$, as defined in Methods. $\Sigma$ merges the messages of the push head with their log-sum-exps $\boldsymbol\psi$, $g$ are the score gates, $\lambda\mathbf c^{\mathrm{pool}}$ is the pooled correction from the tile means $\bar{\mathbf v}_j$ and $\mathbf W^O$ projects the concatenated heads.\par}
\end{minipage}

\clearpage
\par\noindent\begin{minipage}{\textwidth}
\edfig{The noise prior has the spatial and cross-variable structure of ERA5 6\,h residuals.}\label{ed:prior}
\noindent\makebox[\textwidth][c]{\includegraphics[width=180mm]{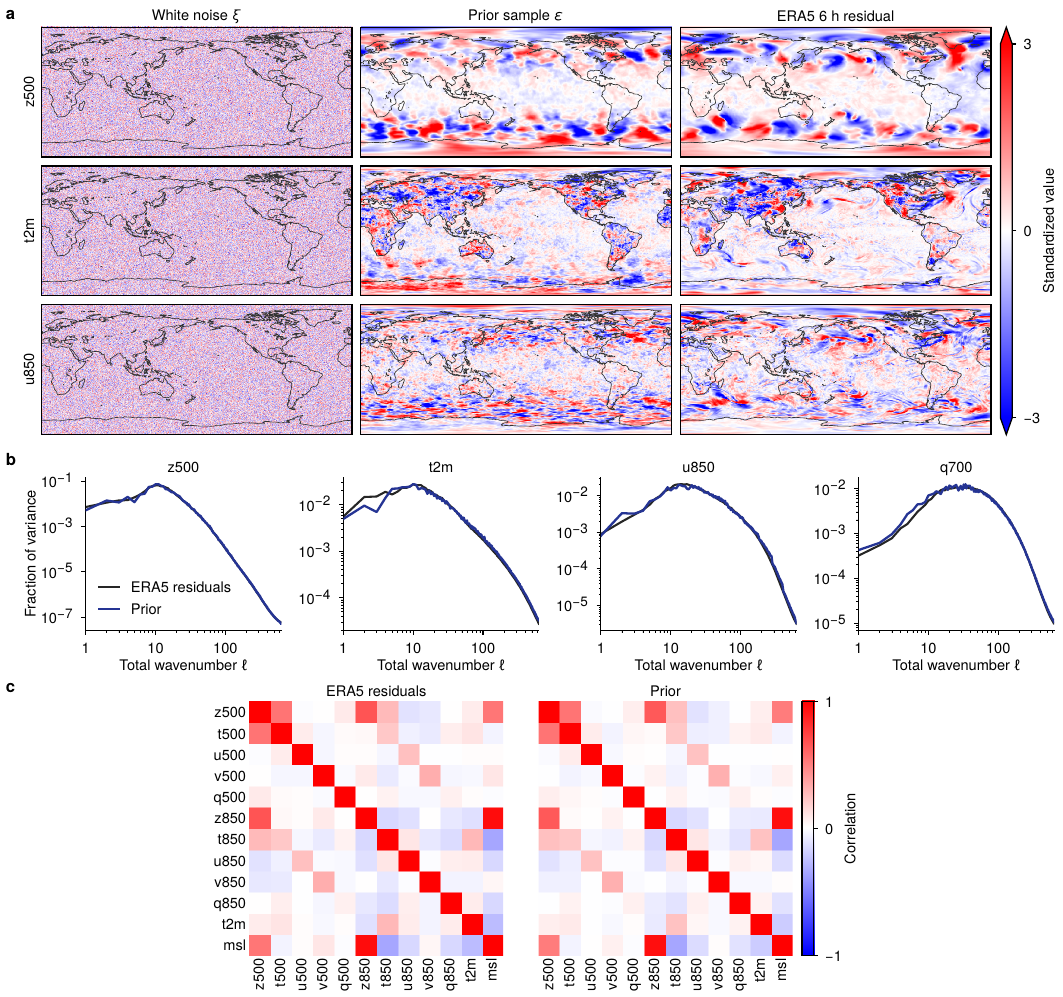}}
\par\smallskip\noindent{\footnotesize
The prior is fitted to ERA5 for 1998--2019; every field shown is from 2020. \textbf{a},~One draw of white noise $\xi$,
the prior sample $\varepsilon$ obtained by colouring it and an ERA5 6\,h residual (00\,UTC, 20 January 2020) for z500,
t2m and u850 (abbreviations as in Fig.~\ref{fig:scorecard}), each divided by its standard deviation. \textbf{b},~Spherical-harmonic power per total wavenumber
$\ell$, normalized to unit total variance, of the prior (8 samples) and of ERA5 residuals (110 pairs at 00\,UTC) for z500, t2m, u850 and q700, up to $\ell = 630$. \textbf{c},~Area-weighted correlations between the residuals
of 12 variables in ERA5 (the 110 pairs of \textbf{b}) and in the prior (its 8 samples); the prior's 66 correlations
between different variables are within 0.07 of those of ERA5 (mean difference 0.01). The ERA5 residuals have their seasonal mean removed, as in the fit (Supplementary Note, section S5).\par}
\end{minipage}
\clearpage
\par\noindent\begin{minipage}{\textwidth}
\edfig{\model's members retain fine-scale spatial power.}\label{ed:spectra}
{\centering\makebox[\textwidth][c]{\includegraphics[width=180mm]{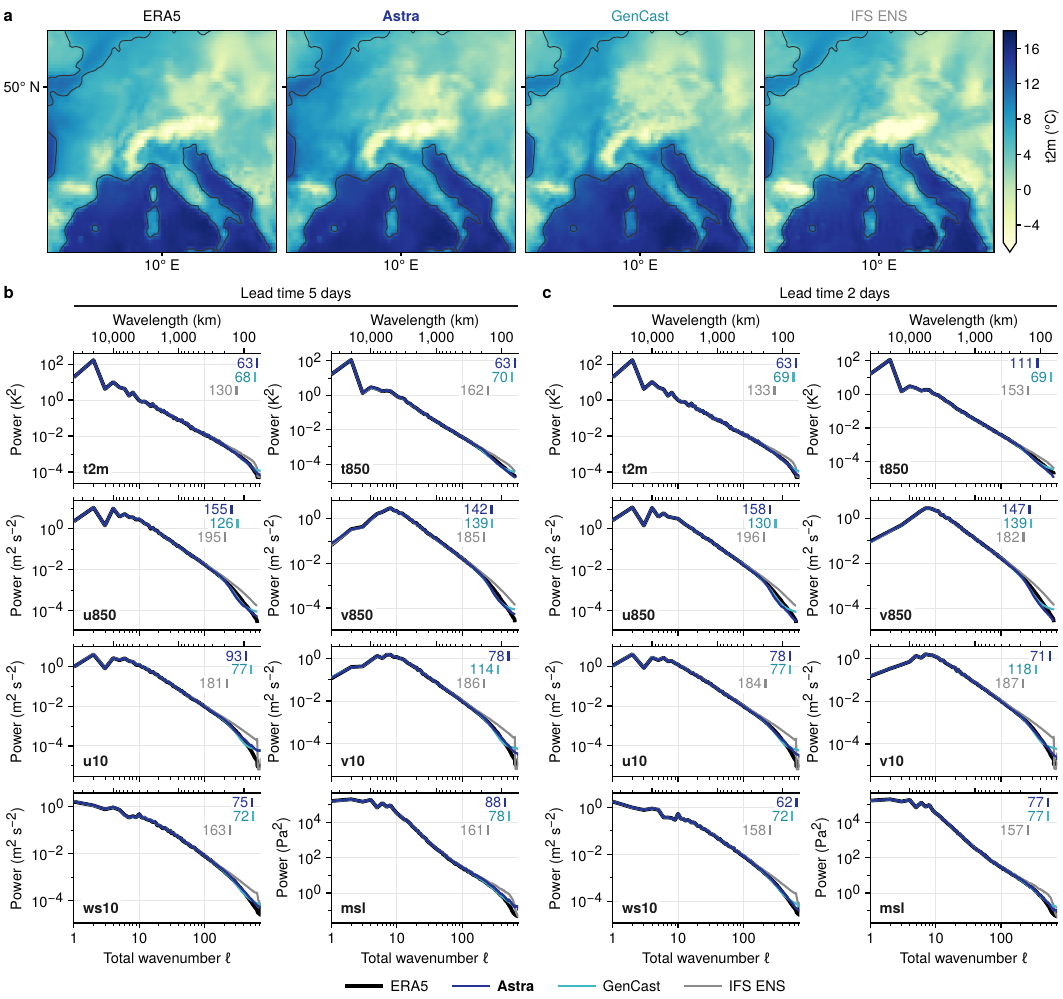}}\par}\vspace{1mm}
\par\noindent{\footnotesize
\textbf{a},~2\,m temperature (t2m) over the Alps (38--54\textdegree\,N, 2\textdegree\,W--22\textdegree\,E), valid at 00\,UTC on 21 December 2020: ERA5 and member~0 of the 5-day forecasts of \model, GenCast and IFS ENS, on one colour scale.
\textbf{b},~Spherical-harmonic power per total wavenumber $\ell$ at a lead time of 5~days, for eight variables (abbreviations as in Fig.~\ref{fig:scorecard}). The top axis gives the wavelength. Coloured ticks, labelled in km, mark the finest wavelength down to which the power stays within a factor of $\sqrt2$ of that of ERA5.
\textbf{c},~As \textbf{b}, at 2~days.
In \textbf{b} and \textbf{c}, \model's ticks lie at shorter wavelengths than those of IFS ENS for all eight variables. Spectra are means over eight members and 24 initializations of 2020. They stop at $\ell=638$ for t850, u850 and v850, below ERA5's TL639 truncation, and at $\ell=719$ for the other variables.\par}
\end{minipage}

\clearpage
\par\noindent\begin{minipage}{\textwidth}
\edfig{Verified against ERA5, \model\ improves on IFS ENS on every target and lead time.}\label{ed:scorecard_full}
{\centering\makebox[\textwidth][c]{\includegraphics[width=180mm]{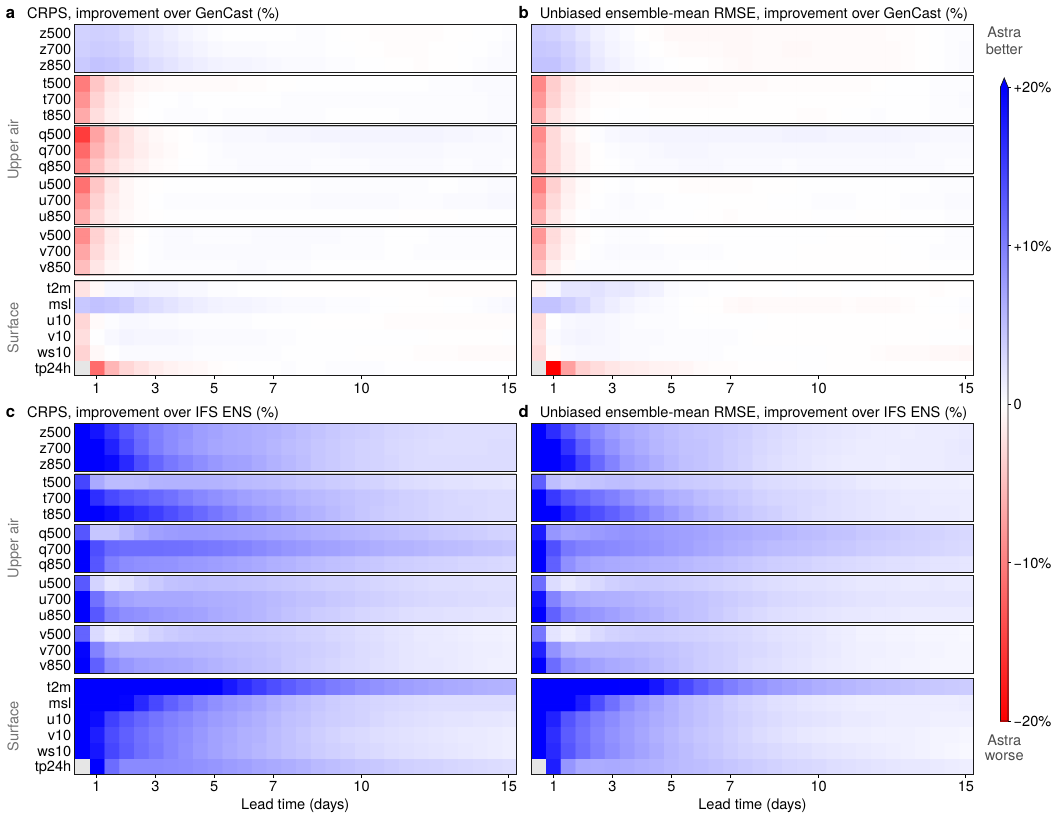}}\par}\vspace{1mm}
\par\noindent{\footnotesize
\textbf{a},~Improvement of \model\ over GenCast in CRPS at 0.25\textdegree, for 21 targets (abbreviations as in Fig.~\ref{fig:scorecard}) at 30 lead times from 12\,h to 15~days. Blue cells show where \model\ is better and red cells where GenCast is, and the colours saturate at 20\%. tp24h is undefined at 12\,h (grey). \model's CRPS is lower on 441 of the 600 combinations of the 20 targets other than tp24h, and on 406 of the 500 at lead times of 3~days or more (81\%).
\textbf{b},~Improvement of \model\ over GenCast in ensemble-mean RMSE. \model's RMSE is lower on 320 of the 600 combinations.
\textbf{c},~Improvement of \model\ over IFS ENS in CRPS. \model's CRPS is lower on all 600 combinations.
\textbf{d},~Improvement of \model\ over IFS ENS in ensemble-mean RMSE. \model's RMSE is lower on all 600 combinations.
All three models are verified against ERA5 over all 732 initializations of 2020.\par}
\end{minipage}

\clearpage
\par\noindent\begin{minipage}{\textwidth}
\edfig{At 1.5\textdegree, \model\ improves on IFS ENS on every target and lead time.}\label{ed:scorecard_1p5}
{\centering\makebox[\textwidth][c]{\includegraphics[width=180mm]{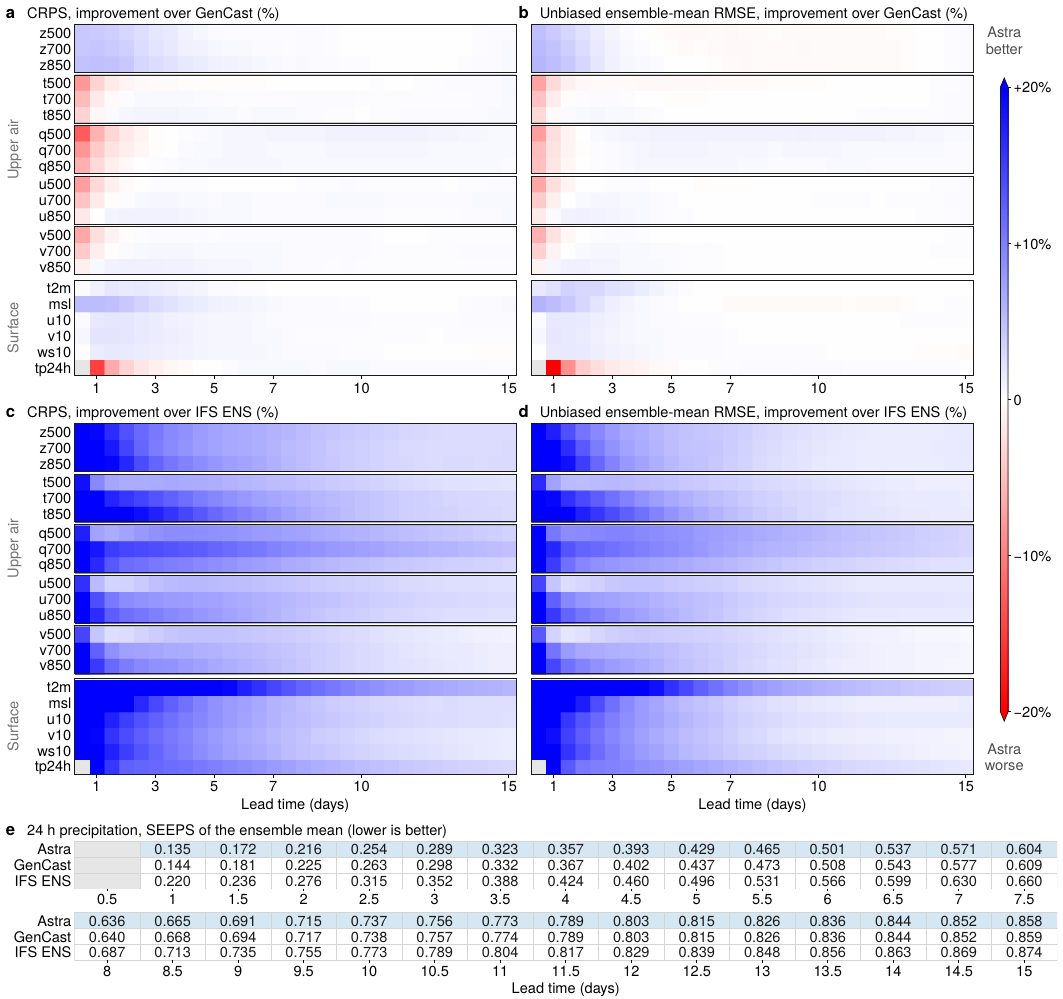}}\par}\vspace{1mm}
\par\noindent{\footnotesize
\textbf{a},~Improvement of \model\ over GenCast in CRPS at 1.5\textdegree, for 21 targets (abbreviations as in Fig.~\ref{fig:scorecard}) at 30 lead times from 12\,h to 15~days. Blue cells show where \model\ is better and red cells where GenCast is, and the colours saturate at 20\%. tp24h is undefined at 12\,h (grey). \model's CRPS is lower on 513 of the 600 combinations of the 20 targets other than tp24h, and on 456 of the 500 at lead times of 3~days or more (91\%).
\textbf{b},~Improvement of \model\ over GenCast in ensemble-mean RMSE. \model's RMSE is lower on 392 of the 600 combinations.
\textbf{c},~Improvement of \model\ over IFS ENS in CRPS. \model's CRPS is lower on all 600 combinations.
\textbf{d},~Improvement of \model\ over IFS ENS in ensemble-mean RMSE. \model's RMSE is lower on all 600 combinations.
\textbf{e},~SEEPS\cite{rodwell2010} of the ensemble mean for tp24h. Shading marks the lowest value before rounding, which is \model's at every lead time.
\model's released forecasts were scored by WeatherBench~2, and the scores of all three models are taken from its result files. All are verified against ERA5 over all 732 initializations of 2020.\par}
\end{minipage}

\clearpage
\par\noindent\begin{minipage}{\textwidth}
\edfig{\model's skill extends across regions.}\label{ed:regions_1p5}
{\centering\makebox[\textwidth][c]{\includegraphics[width=180mm]{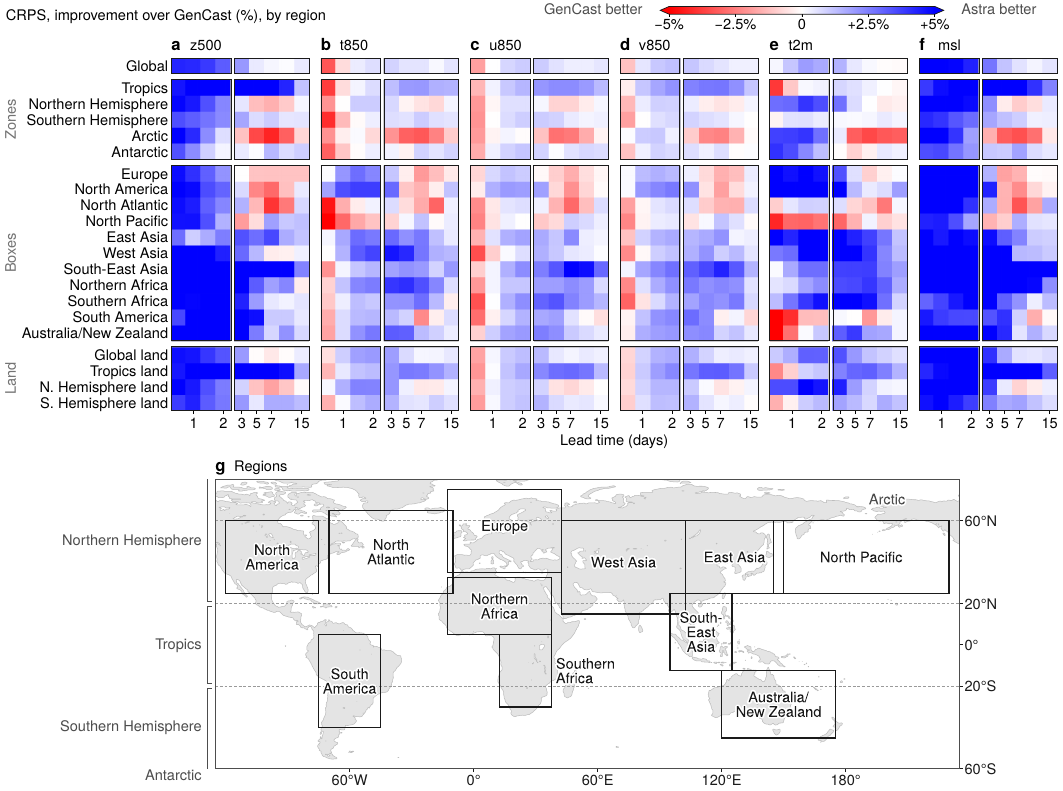}}\par}\vspace{1mm}
\par\noindent{\footnotesize
\textbf{a}--\textbf{f},~Improvement of \model\ over GenCast in CRPS at 1.5\textdegree\ for z500 (\textbf{a}), t850 (\textbf{b}), u850 (\textbf{c}), v850 (\textbf{d}), t2m (\textbf{e}) and msl (\textbf{f}), in 21 regions at nine lead times from 12\,h to 15~days (abbreviations as in Fig.~\ref{fig:scorecard}). Blue cells show where \model\ is better and red cells where GenCast is, and the colours saturate at 5\%. \model's CRPS is lower in 873 of the 1,134 combinations of region, variable and lead time (77\%), and in every region for z500 and msl over the first 2~days. In the same evaluation, \model's CRPS is lower than that of IFS ENS in 1,119 combinations.
\textbf{g},~The regions, which are those of WeatherBench-X: the tropics (20\textdegree\,S--20\textdegree\,N), the two hemispheres poleward of 20\textdegree, the Arctic and Antarctic poleward of 60\textdegree, the 11 boxes drawn here, and the land points of the global, tropical and hemispheric regions.
Scores are from the evaluation of Extended Data Fig.~\ref{ed:scorecard_1p5}.\par}
\end{minipage}

\clearpage
\par\noindent\begin{minipage}{\textwidth}
\edfig{\model's skill holds for each variable and its spread matches its error at 15~days.}\label{ed:curves}
{\centering\makebox[\textwidth][c]{\includegraphics[width=180mm]{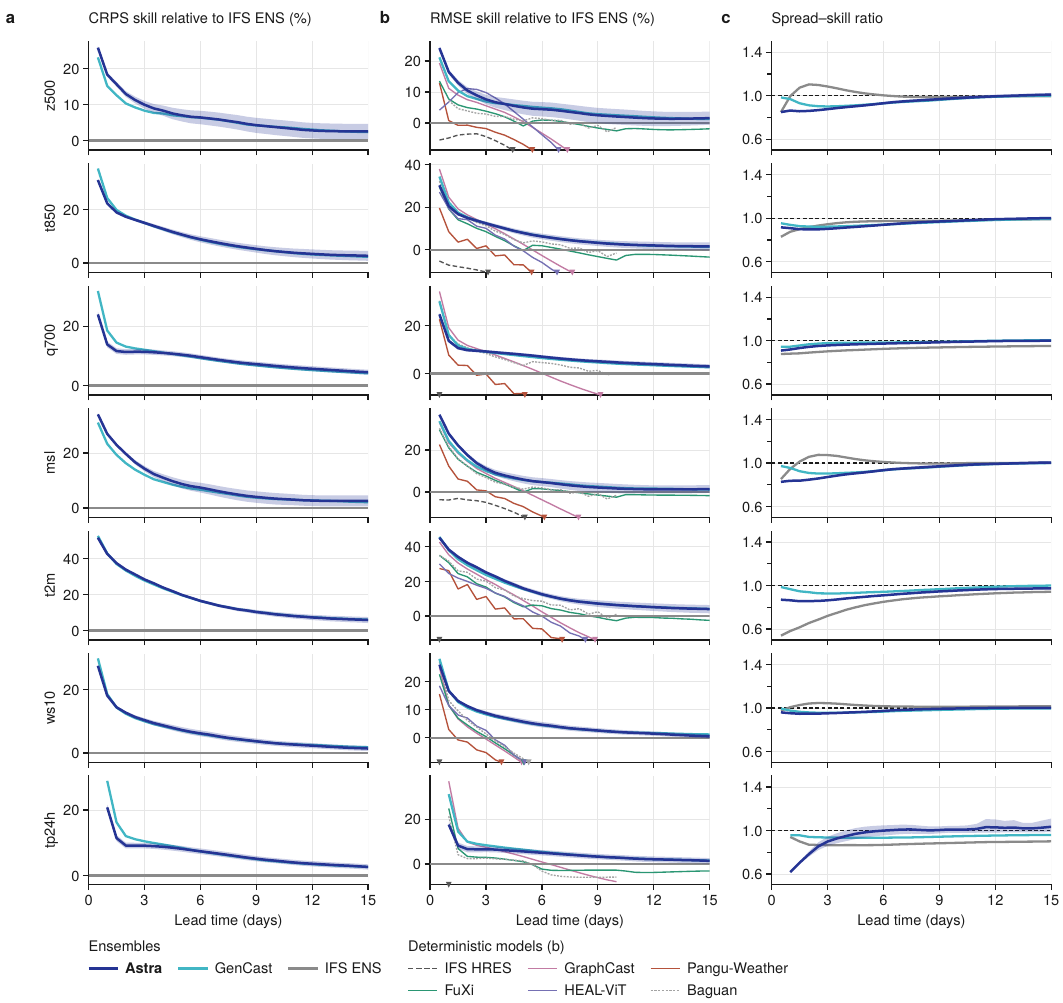}}\par}\vspace{1mm}
\par\noindent{\footnotesize
\textbf{a},~CRPS skill of \model\ and GenCast relative to IFS ENS, for seven variables at lead times up to 15~days (abbreviations as in Fig.~\ref{fig:scorecard}). Skill is the percentage by which a score is lower than that of IFS ENS, drawn as the line at zero. \model's skill is positive for every variable at every lead time.
\textbf{b},~Ensemble-mean RMSE skill relative to IFS ENS, for the same variables. \model's skill is again positive throughout. The panel also shows six deterministic models: IFS HRES\cite{haiden2021}, GraphCast\cite{lam2023}, FuXi\cite{chen2023}, Pangu-Weather\cite{bi2023}, \mbox{HEAL-ViT}\cite{ramavajjala2024} and Baguan\cite{niu2025}. Each is drawn for the variables that WeatherBench~2 scores for it, up to its last lead time. A curve that falls below a panel is cut at its lower edge and marked by a triangle.
\textbf{c},~Spread--skill ratio of the three ensembles. The dashed line at 1 marks a spread equal to the ensemble-mean error. At 15~days, \model's ratio lies between 0.97 and 1.03 for all seven variables.
Shading shows 95\% block-bootstrap confidence intervals for \model. All models are verified against ERA5 at 0.25\textdegree. \model, GenCast and IFS ENS are scored over all 732 initializations of 2020.\par}
\end{minipage}

\clearpage
\par\noindent\begin{minipage}{\textwidth}
\edfig{Two-sided routing recovers optimal pair selections, and learning the two selections benefits from independent scores.}\label{ed:routing_toy}
{\centering\makebox[\textwidth][c]{\includegraphics[width=180mm]{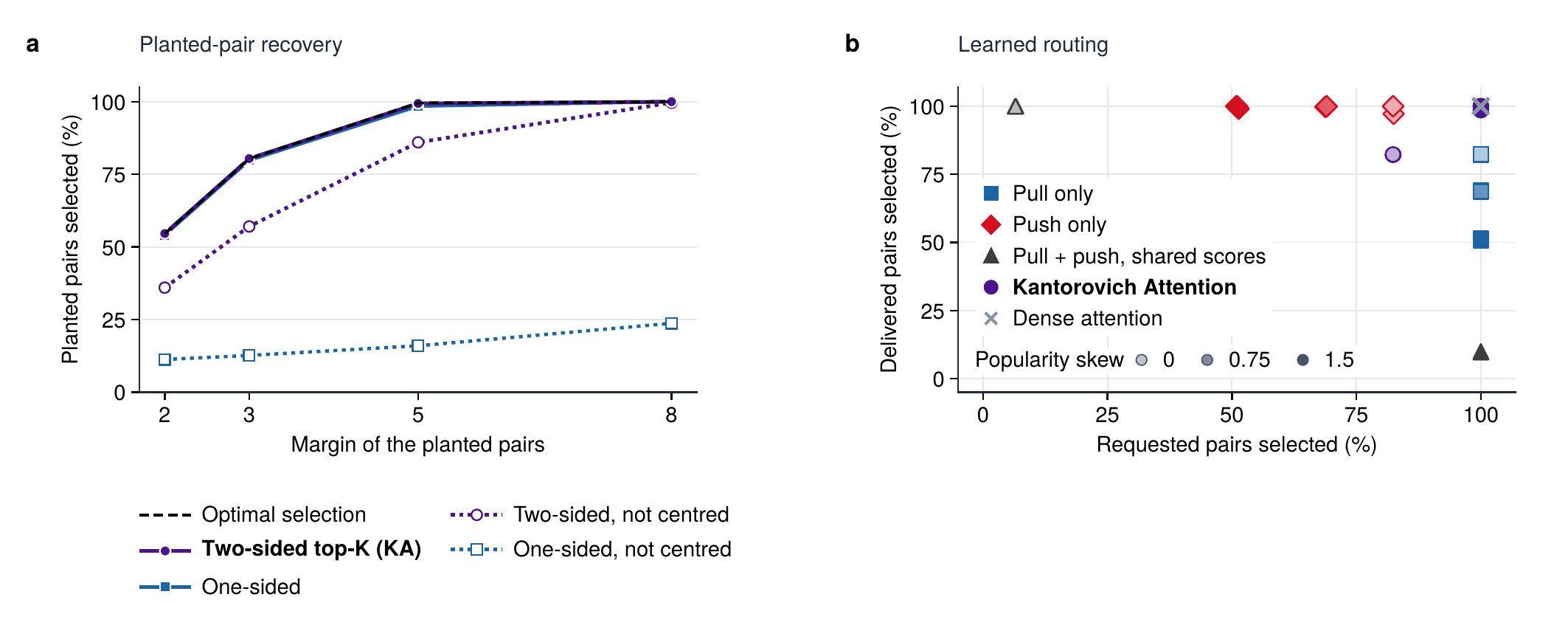}}\par}\vspace{1mm}
\par\noindent{\footnotesize
\textbf{a},~Recovery of planted pairs from synthetic scores on 180 tiles. Each tile is planted to receive from 8 tiles and to send to 8 ($K=8$). A score is the margin times the planted indicator plus standard normal noise, a standard normal bias for its receiver and, for its sender, twice a squared standard exponential draw. Lines show the share of planted pairs selected by two-sided top-$K$, by one-sided selection (the largest scores of each receiver) and by the optimal selection, which is the maximum-score selection with at least $K$ senders and $K$ receivers per tile (Supplementary equation~\eqref{eq:lp}). All select the same number of pairs, and each point is the mean of five score matrices. Dotted lines omit centring. With centring, two-sided top-$K$ captures 99.8--100\% of the optimal selection's score gain over random pairs.
\textbf{b},~Learned routing with one layer of two-branch attention on 64 tiles ($K=4$). An input names, for each tile, 4 tiles to receive from (requests) and 4 receivers for its tokens (deliveries). How often a tile is named follows a log-normal distribution with spread 0, 0.75 or 1.5 (skew). Points show the recall of requested and of delivered pairs for 3 seeds per skew. The variants have the same model size and training settings: pull only (two pull branches), push only (two push branches), pull and push with shared routing scores, \hla\ with independent routing scores, and dense attention. Every receiver also attends to its own tile.\par}
\end{minipage}

\clearpage
\par\noindent\begin{minipage}{\textwidth}
\edtable{Network, state variables and training.}\label{edt:cost}
{\footnotesize\noindent\textbf{a},~Parameters and forward cost by level.\par\smallskip}
{\footnotesize\setlength{\tabcolsep}{5pt}\noindent\begin{tabular*}{\textwidth}{@{\extracolsep{\fill}}lllllrr@{}}
\toprule
Level & Grid & Width & Blocks & Attention & Parameters & TFLOP/sample\\
\midrule
L0 & $721\times1{,}440$ & 96 & 4 & -- (convolution) & 2\,M & 1.53\\
L1 & $181\times360$ & 384 & 16 & \hla, $8\times8$ tiles & 52\,M & 7.10\\
L2 & $91\times180$ & 768 & 24 & \hla, $10\times10$ tiles & 245\,M & 10.23\\
L3 & $46\times90$ & 1,920 & 16 & dense & 718\,M & 7.98\\
L4 & $23\times45$ & 3,072 & 16 & dense & 1,828\,M & 3.98\\
\midrule
Network & & & & & 3.0\,B & 32.0\\
\bottomrule
\end{tabular*}\par\smallskip
\footnotesize\noindent Forward floating-point operations per sample, two per multiply--add, from the traced program of the trained network. The network totals also include the resamplers, the conditioning, the embeddings and the output head.\par}
\par\smallskip
{\footnotesize\noindent\textbf{b},~Variables and normalization.\label{edt:vars}\par\smallskip}
{\footnotesize\setlength{\tabcolsep}{4pt}\renewcommand{\arraystretch}{0.95}\noindent\begin{tabular*}{\textwidth}{@{\extracolsep{\fill}}L{0.44\textwidth}L{0.17\textwidth}L{0.19\textwidth}l@{}}
\toprule
Variables & Domain & Transform & Noised\\
\midrule
Geopotential, temperature, specific humidity, $u$ and $v$ wind & global & -- & yes\\
2\,m temperature, 10\,m $u$ and $v$ wind, mean sea-level pressure & global & -- & yes\\
6\,h accumulated precipitation & global & $\log(1 + x/10^{-5}\,\mathrm{m})$ & yes\\
Sea-surface temperature, sea-ice cover & ocean & -- & no\\
Volumetric soil water (layers 1 and 2) & land & -- & no\\
Snow depth & land & $\log(1 + x/10^{-3}\,\mathrm{m})$ & no\\
\bottomrule
\end{tabular*}\par\smallskip
\footnotesize\noindent The 75 state variables. The five variables of the first row are on 13 pressure levels (50, 100, 150, 200, 250, 300, 400, 500, 600, 700, 850, 925 and 1,000\,hPa) and the others on a single level. Each variable, after its transform, is standardized with its mean and standard deviation, and its
6\,h residual is divided by its standard deviation $\boldsymbol{\sigma}_\Delta$. Ocean and land fields are zero outside their domain. Static fields and forcings are listed in Supplementary Note, section S5.\par}
\par\smallskip
{\footnotesize\noindent\textbf{c},~Training.\label{edt:hyp}\par\smallskip}
{\footnotesize\setlength{\tabcolsep}{4pt}\renewcommand{\arraystretch}{0.93}\noindent\begin{tabular*}{\textwidth}{@{\extracolsep{\fill}}L{0.16\textwidth}L{0.37\textwidth}L{0.42\textwidth}@{}}
\toprule
 & MeanFlow training & SF-CRPS training\\
\midrule
Initialization & random & from MeanFlow training; new optimizer state\\
Updates; batch & 40,000; 16 samples & 15,000; 8 initializations $\times$ 4 members\\
Objective & MeanFlow & SF-CRPS: afCRPS on 5 scale bands (weights 0.10, 0.25, 0.25, 0.20, 0.20, finest first) and the feature term\\
Optimizer & schedule-free Muown ($\beta_1 = 0.9$, momentum 0.95, update RMS 0.2) & as in the first stage\\
Learning rate & $4\times10^{-4}$ & $4\times10^{-5}$ ($10^{-4}$ for the adapters)\\
Gradient clipping (norm) & 8 & 5\\
Rollout & one 6\,h step & two 6\,h steps; lead weights 0.25 and 0.75\\
Feature term & -- & weight 0.5; decoder~1 at L1--L4 (weights 0.4, 0.2, 0.2, 0.2); steps for 12, 18, 24 and 30\,h (probabilities 0.35, 0.35, 0.20, 0.10); $\tau \in [0.5, 1)$, $r \in [0, \tau)$; copy refreshed every 5,000 updates\\
Noisy adapter & -- & rank 16, at L3 and L4\\
Time per update; compute & 1.94\,s; 173 H200-hours & 8.29\,s; 276.3 H200-hours\\
\bottomrule
\end{tabular*}\par\smallskip
\noindent Both stages: variable weights pressure/1,000\,hPa (1 for single-level fields) $\times$ 1.5 (q), 1.3 (t, u, v, 10\,m wind), 1.35 (z), 1.8 (msl), 8.0 for 2\,m temperature, 0.66 for 6\,h precipitation and sea-surface temperature and 0.33 for the other auxiliary fields; constant learning rate after 500 warm-up updates; no weight decay.\par}
\end{minipage}

\clearpage
\par\noindent\begin{minipage}{\textwidth}
\edtable{What one block of \hla\ can hold, for suitable scores.}\label{edt:ka_hold}
{\footnotesize\setlength{\tabcolsep}{2pt}\noindent\begin{tabular*}{\textwidth}{@{\extracolsep{\fill}}lllll@{}}
\toprule
 & \multicolumn{2}{l}{1\textdegree\ level (L1)} & \multicolumn{2}{l}{2\textdegree\ level (L2)}\\
\cmidrule(lr){2-3}\cmidrule(l){4-5}
Quantity & Theorem & \model & Theorem & \model\\
\midrule
Selections per tile and side & $K\ge1$ & 8 & $K\ge1$ & 8\\
Head pairs per block & $h$ & 3 & $h$ & 6\\
Tiles, for one head pair & $T\ge2K+2=18$ & 1,035$^{*}$ & $T\ge2K+2=18$ & 180\\
Tiles, for one block & $T\ge2hK+2=50$ & 1,035$^{*}$ & $T\ge2hK+2=98$ & 180\\
Descriptor width & $d_k\ge2K+2=18$ & 128 & $d_k\ge2K+2=18$ & 128\\
Input (level) width & $C\ge h(8K+8)=216$ & 384 & $C\ge h(8K+8)=432$ & 768\\
Selections per head pair & $2TK$ & 16,560 & $2TK$ & 2,880\\
Fewest pairs for any $K$ problems & $K(2T-1)$ & 16,552 & $K(2T-1)$ & 2,872\\
Share of the $T(T-1)$ tile pairs & $K(2T-1)/T(T-1)$ & 1.55\% & $K(2T-1)/T(T-1)$ & 8.91\%\\
Problems one head pair can hold & exactly $K$ & 8 & exactly $K$ & 8\\
Problems one block can hold & exactly $hK$ & 24 & exactly $hK$ & 48\\
\midrule
Units at unit cost, one side$^{\dagger}$ & $\ge T{-}1{-}2K$ of $2T{-}1$ & $\ge$1,018 of 2,069 & $\ge T{-}1{-}2K$ of $2T{-}1$ & $\ge$163 of 359\\
\bottomrule
\end{tabular*}\par\smallskip
\footnotesize\noindent A set of pairs holds a problem when, for some KA inputs and parameters chosen for the problems,
the self pairs and the selected pairs together contain an optimal plan of it; the selected pairs of a block are
the union of those of its head pairs. ``Exactly'' means every collection of that many problems and not every
collection of one more (Supplementary Theorem~\ref{thm:ka_support} and Proposition~\ref{prop:ka_budget}). Theorem columns give the condition or count, and \model\ columns give the values of its configuration (Methods and
Extended Data Table~\ref{edt:cost}a). The fewest pairs, $K(2T-1)$, are needed
and suffice for $T\ge2K+1=17$; one head pair needs only $C\ge8K+8=72$; none of the conditions on $T$ can be lowered.
The $T(T-1)$ tile pairs are the pairs of distinct tiles. $^{\dagger}$Not \model's selection: for one problem with optimal cost 0 and
$2T-1$ units of mass, every plan contained in the pairs of two heads that both select at the receiver, or both at the
sender, moves at least this many units at unit cost (Supplementary Theorem~\ref{thm:ka_support}(ii) with $m=2$), shown at \model's $T$ and $K$.
$^{*}$Alternate blocks shift the tiles at 1\textdegree: $T=1{,}080$, $2TK=17{,}280$, $K(2T-1)=17{,}272$, a share of
1.48\% and, with both heads on one side, at least 1,063 of 2,159 units.\par}
\end{minipage}

\clearpage
\section{Supplementary Note}
\setcounter{equation}{0}\renewcommand{\theequation}{S\arabic{equation}}%
\setcounter{proposition}{0}\renewcommand{\theproposition}{S\arabic{proposition}}%
\setcounter{theorem}{0}\renewcommand{\thetheorem}{S\arabic{theorem}}%
Sections S1--S3 cover \hla, S4--S6 the network, the data and training, S7 costs and S8--S9 forecast evaluation. S1 includes Theorem~\ref{thm:ka_support} and Propositions~\ref{prop:ka_budget} and~\ref{prop:ka_shared} with their proofs.

\subsection{S1. Attention operator}
\subsubsection*{Operator specification}
The input is $\mathbf X=\mathrm{Derf}(\mathbf h+\mathbf e)$, with one learned $P^2$-vector position table $\mathbf e$ per level and $h$ heads per branch (3 at L1, 6 at L2). Pull queries are scaled channel-wise by $1+\mathbf a_i$, learned per tile and initially zero. With selected-token outputs $\mathbf Y_i^{b,h}$ and pooled vectors $\mathbf c_i^{\mathrm{pool},b,h}$ (equation~\eqref{eq:tail}) for branch $b$ and head $h$,
\begin{equation}
\mathrm{KA}(\mathbf{X})_i =
\Bigl[\;\mathop{\Big\Vert}_{b\in\{\mathrm{pull},\,\mathrm{push}\}}\;\mathop{\Big\Vert}_{h=1}^{H}
\bigl(\mathbf{Y}^{b,h}_i + \lambda^{b,h}\,\mathbf{1}\,\bigl(\mathbf{c}_i^{\mathrm{pool},b,h}\bigr)^\top\bigr)\Bigr]\,
\mathbf{W}^{O},
\label{eq:ha}
\end{equation}
where $\mathbf Q_i,\mathbf K_i,\mathbf V_i\in\mathbb R^{P^2\times d_k}$ hold the token rows of tile $i$, $\mathbf W^O\in\mathbb R^{2Hd_k\times C}$, $\Vert$ concatenates along channels, $\mathbf{1} \in \mathbb{R}^{P^2}$, and $\lambda^{b,h} \in (0,1)$ is sigmoid-parametrized ($1/4$ at initialization). Below, $b$ and $h$ are dropped. Tiles wrap in longitude, and padded tokens beyond the poles are excluded from every pooling, mean and softmax.

\paragraph{Tile scores and selection.} Descriptors are pooled from the tokens of each tile before any rotary rotation,
\begin{equation}
\hat{\mathbf{q}}_i = \rho\bigl(\mathrm{pool}_{\mathbf{w}_Q}(\mathbf{Q}_i)\bigr), \qquad
\hat{\mathbf{k}}_i = \rho\bigl(\mathrm{pool}_{\mathbf{w}_K}(\mathbf{K}_i)\bigr), \qquad
\mathrm{pool}_{\mathbf{w}}(\mathbf{Z}_i) = \sum_{p \in i}
  \frac{e^{\mathbf{w}^{\top}\mathbf{z}_p}}{\sum_{p' \in i} e^{\mathbf{w}^{\top}\mathbf{z}_{p'}}}\,\mathbf{z}_p ,
\label{eq:nodes}
\end{equation}
where the pooling vectors start at zero (uniform tile means) and $\rho$ subtracts the unweighted mean across tiles and scales each descriptor to unit RMS. Scores are $s_{ij}=\hat{\mathbf q}_i^\top\hat{\mathbf k}_j/\sqrt{d_k}$, i.e.\ $\sqrt{d_k}$ times centred cosine similarity. Branches and heads score and select independently; same-index pull and push heads form a head pair (Fig.~\ref{fig:ka}a). Tile descriptors use learned token pooling\cite{wu2024transolver} only to rank tiles. For unnormalized descriptors $\mathbf q_i,\mathbf k_j$ with tile means $\bar{\mathbf q},\bar{\mathbf k}$,
\begin{equation}
\mathbf{q}_i^{\top}\mathbf{k}_j = (\mathbf{q}_i-\bar{\mathbf{q}})^{\top}(\mathbf{k}_j-\bar{\mathbf{k}}) + \bar{\mathbf{q}}^{\top}(\mathbf{k}_j-\bar{\mathbf{k}}) + (\mathbf{q}_i-\bar{\mathbf{q}})^{\top}\bar{\mathbf{k}} + \bar{\mathbf{q}}^{\top}\bar{\mathbf{k}};
\label{eq:centring}
\end{equation}
centring removes the sender-only and receiver-only terms. The selected sets $\mathcal N(i)$, $\mathcal R(j)$ and $\mathcal S(i)$, each including the tile itself, are those of equation~\eqref{eq:graph}. Both selections use a fixed tie-breaking rule.
Let $S^{\mathrm{pull}}$ and $S^{\mathrm{push}}$ be the score matrices of a head pair, indexed by receiver $i$ and sender $j$, and let $\operatorname{RowTop}_K$ and $\operatorname{ColTop}_K$ mark the $K$ largest off-diagonal scores of each row and of each column. For the off-diagonal masks $A^{\mathrm{pull}}=\operatorname{RowTop}_K(S^{\mathrm{pull}})$ and $A^{\mathrm{push}}=\operatorname{ColTop}_K(S^{\mathrm{push}})$, the selected pairs $\Pi^\circ=A^{\mathrm{pull}}\cup A^{\mathrm{push}}$ satisfy
\begin{equation}
\begin{aligned}
A^{\mathrm{pull}}\mathbf{1}&=K\mathbf{1},\qquad
(A^{\mathrm{push}})^{\top}\mathbf{1}=K\mathbf{1},\\
|\Pi^{\circ}|&=2TK-|A^{\mathrm{pull}}\cap A^{\mathrm{push}}|,
\qquad TK\le|\Pi^{\circ}|\le 2TK.
\end{aligned}
\label{eq:ka_support_count}
\end{equation}

\paragraph{Attention over selected tiles.} Queries and keys carry spherical rotary embeddings\cite{su2024}, applied to the queries after the flow-time phase (section S4). Channel pair $n$ of the token at unit position vector $\mathbf{p}$ is rotated by $\pi\,\mathbf{p}^{\top}\mathbf{d}_n$, with $d_k/2 = 64$ unit directions $\mathbf{d}_n$ on a Fibonacci lattice, shared by all heads; the query--key product then depends on position only through the chord $\mathbf{p} - \mathbf{p}'$, continuously across the date line and poles. With $\bar s_i$ the mean score of the senders of $i$ (its own tile included), the score gate of equation~\eqref{eq:transport} is
\begin{equation*}
g(s_{ij}) = 1 + \gamma_g\,\mathrm{erf}\bigl(\omega_g(s_{ij} - \bar s_i) +
\delta_g\bigr),\qquad \omega_g>0,
\end{equation*}
of the form of equation~\eqref{eq:derf}. At initialization $\gamma_g = 1$ and $\delta_g = 0$, so the gate lies in $(0,2)$ and equals 1 for a sender of mean score; $\omega_g$ is 0.94 at L1 and 1.06 at L2 in pull, and 1.06 and 1.20 in push. In equation~\eqref{eq:transport}, $\mathbf{Q}$ and $\mathbf{K}$ are taken after the rotations, $\mathbf{K}_{\Pi(i)},\mathbf{V}_{\Pi(i)}$ stack the tokens of the tiles in $\Pi(i)=\{j:\Pi_{ij}=1\}$, and $\mathbf{G}_i$ holds $g(s_{ij})$ on the rows of tile $j$. In pull, $\Pi^{\mathrm{pull}}_{ij}=[\,j\in\mathcal N(i)\,]$ and each receiver evaluates one softmax\cite{vaswani2017} over the $(K+1)P^2$ tokens of its senders. In push, every sender $j$ forms one message for each selected receiver $i\in\mathcal R(j)$, in which the receiver's queries attend to the sender's tokens, and keeps the per-query log-sum-exp $\boldsymbol{\psi}_{j\to i} \in \mathbb{R}^{P^2}$,
\begin{equation}
\mathbf{Y}_{j\to i} = \operatorname{softmax}\!\Bigl(
  \frac{\mathbf{Q}_i \mathbf{K}_j^{\top}}{\sqrt{d_k}} \Bigr) \mathbf{V}_j,
\qquad
\boldsymbol{\psi}_{j\to i} = \operatorname{logsumexp}\Bigl(
  \frac{\mathbf{Q}_i \mathbf{K}_j^{\top}}{\sqrt{d_k}} \Bigr),
\qquad i \in \mathcal{R}(j).
\label{eq:msg}
\end{equation}
Each receiver merges its messages token by token (Extended Data Fig.~\ref{ed:arch}e),
\begin{equation}
\mathbf{Y}^{\mathrm{push}}_i = \sum_{j \in \mathcal{S}(i)} g(s_{ij})\,
  \frac{\exp \boldsymbol{\psi}_{j\to i}}{\sum_{j'\in\mathcal{S}(i)} \exp \boldsymbol{\psi}_{j'\to i}}
  \odot \mathbf{Y}_{j\to i}
  = \operatorname{softmax}\!\Bigl(
  \frac{\mathbf{Q}_i \mathbf{K}_{\mathcal{S}(i)}^{\top}}{\sqrt{d_k}} \Bigr)
  \bigl( \mathbf{G}_i \odot \mathbf{V}_{\mathcal{S}(i)} \bigr),
\label{eq:push}
\end{equation}
which is equation~\eqref{eq:transport} with $\Pi(i)=\mathcal S(i)$; in a pull head, $\Pi(i)=\mathcal N(i)$. This uses $T(K+1)$ blocks of $P^2\times P^2$ token pairs per head, plus merging, however messages spread over receivers. Unlike bi-level routing\cite{zhu2023biformer}, KA selects at the sender as well as the receiver and adds score centring, score gates and the pooled correction. Unlike Sinkformers\cite{sander2022}, which make dense attention doubly stochastic by Sinkhorn normalization, KA selects on both sides and keeps the softmax of each receiver.

\paragraph{Pooled term.} Tiles that receiver $i$ does not select, $\Pi(i)^{\mathrm c}$, contribute through the means of their values. The pooled correction is
\begin{equation}
\mathbf{c}_i^{\mathrm{pool}} = \langle \bar{\mathbf{v}}\rangle_{\Pi(i)^{\mathrm{c}}} -
  \langle \bar{\mathbf{v}}\rangle_{\Pi(i)},
\qquad
\langle \bar{\mathbf{v}}\rangle_{\mathcal{A}} = \sum_{j\in \mathcal{A}}
  \pi_{\mathcal{A}}(j\mid i)\,\bar{\mathbf{v}}_j,
\qquad
\pi_{\mathcal{A}}(j\mid i) = \frac{e^{\kappa s_{ij}}}{\sum_{j'\in \mathcal{A}} e^{\kappa s_{ij'}}},
\label{eq:tail}
\end{equation}
where $\bar{\mathbf v}_j$ is the mean of $\mathbf V_j$ over the tokens of tile $j$. The weights $\pi$ use the routing scores of the same head with a learned inverse temperature $\kappa\in(0,1]$ that starts at $1/2$. Subtracting the selected-tile mean avoids double-counting content already carried by the selected tokens, and the correction vanishes when all tile means agree. The correction is added to each token of the receiver with a learned weight $\lambda\in(0,1)$. Because the softmaxes of equation~\eqref{eq:tail} cover all tiles, every score, selected or not, receives a derivative proportional to $\kappa\,\pi_{\mathcal A}(j\mid i)(\bar{\mathbf v}_j-\langle\bar{\mathbf v}\rangle_{\mathcal A})$ (negative for selected tiles), so a tile outside the top $K$ can rise into it. The pooling vectors, $\kappa$, $\lambda$ and the gate parameters $(\gamma_g,\omega_g,\delta_g)$ are learned per head and branch. The gate parameters also depend on the flow times $(r,\tau)$: each is a learned value plus a linear function, zero at initialization, of the flow-time features of equation~\eqref{eq:derf}, passed through a softplus for $\omega_g$.

\subsubsection*{Theorem~\ref{thm:ka_support} and Propositions~\ref{prop:ka_budget} and~\ref{prop:ka_shared} with their proofs}
\makeatletter
\providecommand{\proofrhythm}{%
  \setlength{\abovedisplayskip}{10pt plus 3pt minus 2pt}%
  \setlength{\belowdisplayskip}{10pt plus 3pt minus 2pt}%
  \setlength{\abovedisplayshortskip}{6pt plus 2pt minus 1pt}%
  \setlength{\belowdisplayshortskip}{8pt plus 2pt minus 2pt}%
  \setlength{\jot}{5pt}}
\providecommand{\proofstyle}{%
  \let\proofstyle@normalsize\normalsize
  \renewcommand{\normalsize}{\proofstyle@normalsize\proofrhythm}%
  \proofrhythm}
\makeatother
\newenvironment{proofblock}[1][Proof]%
  {\par\addvspace{\medskipamount}\noindent\textit{#1.}\enspace\ignorespaces}%
  {\nobreak\hspace*{\fill}\nobreak\enspace$\square$\par\addvspace{\bigskipamount}}
\newcounter{suppfig}
\newcommand{\suppfig}[1]{%
  \refstepcounter{suppfig}%
  \par\medskip\noindent{\small\textbf{Supplementary Fig.\
  \arabic{suppfig}\,$\vert$\,#1}}\par\smallskip}

\begingroup\proofstyle
Let $I$ be the set of self pairs $(i,i)$, and write $E=\{(i,j):i\ne j\}$ for the non-self pairs,
the edges between $T$ receivers $i$ and $T$ senders $j$. Problem $\ell$ has a finite cost matrix $D_\ell$ and masses
$a_\ell,b_\ell>0$ of equal total mass. Its plans, and its least cost over a set $\mathcal G$ of pairs, are
\begin{equation*}
\mathcal U_\ell=\{\Phi\ge0:\Phi\mathbf1=a_\ell,\ \Phi^\top\mathbf1=b_\ell\},\qquad
\mathcal V_\ell(\mathcal G)=\min_{\substack{\Phi\in\mathcal U_\ell\\\operatorname{supp}\Phi\subseteq\mathcal G}}
\langle D_\ell,\Phi\rangle.
\end{equation*}
Here $\operatorname{supp}\Phi=\{(i,j):\Phi_{ij}>0\}$ is the support of the plan $\Phi$, the pairs along which it moves
mass, so $\mathcal G$ contains $\Phi$ when $\operatorname{supp}\Phi\subseteq\mathcal G$. An infeasible restricted
minimum is $+\infty$, and $\mathcal V_\ell:=\mathcal V_\ell(I\cup E)$ is the unrestricted minimum. A set $\mathcal G$ of non-self pairs holds a collection of problems, or supports an optimal plan for each of them, if $\mathcal V_\ell(I\cup\mathcal G)=\mathcal V_\ell$ for each. Throughout, $K\ge1$ and every tile holds at least one grid point.

\begin{theorem}[Optimal plans in one head pair]\label{thm:ka_support}
\noindent\textbf{(i) Selection on both sides.}
Let $T\ge2K+1$, $d_k\ge2K+2$ and input width $C\ge8K+8$. For every collection of $K$ linear transport problems there are KA inputs and parameters, depending on the collection, such that one KA head pair has $K(2T-1)$ selected pairs, which hold the collection:
\begin{equation}
\mathcal V_\ell(I\cup\Pi^\circ)=\mathcal V_\ell\qquad(\ell=1,\ldots,K).
\label{eq:ka_optimal_support}
\end{equation}
Restricted to these pairs, transport thus costs the same as over all pairs.

\noindent\textbf{(ii) Selection on one side.}
Let $m$ heads all select at the receiver, or all at the sender, $K$ each, and let $T\ge mK+2$. There is a linear transport problem with costs 0 and 1 and optimal cost 0 such that the following holds for either rule and for any scores. Every plan contained in the self pairs and the selected pairs of the heads moves at least $T-1-mK$ of its $2T-1$ units of mass at unit cost.
\end{theorem}

A \emph{row-$K$ mask} is a $T\times T$ matrix of zeros and ones with $K$ ones in every row, and a \emph{column-$K$
mask} one with $K$ ones in every column. For a mask $A$, write $A(i)=\{j:A_{ij}=1\}$ for the senders that row $i$
selects. A mask is \emph{non-self}, or \emph{zero-diagonal}, if $i\notin A(i)$ for every $i$; its ones are then pairs
of $E$. The masks $A^{\mathrm{pull}}$ and $A^{\mathrm{push}}$ of equation~\eqref{eq:ka_support_count} are a non-self
row-$K$ mask and a non-self column-$K$ mask, and a head pair selects the pairs
$\Pi^\circ=A^{\mathrm{pull}}\cup A^{\mathrm{push}}$.

\begin{lemma}[Normalized descriptor realization]\label{lemma:ka_realisation}
Let $T\ge3$ and let $A$ be a zero-diagonal row-$K$ mask. There exist $q_i,k_j\in\mathbb R^{2K+2}$, a radius $R>0$
and constants $c_i$ such that
\begin{gather*}
\sum_i q_i=\sum_j k_j=0,\qquad q_i\ne0,\qquad \|k_j\|=R,\\
q_i^\top k_j=c_i\quad\text{for } j\in A(i),\qquad q_i^\top k_j<c_i\quad\text{otherwise}.
\end{gather*}
\end{lemma}

\begin{proofblock}
The first $2K$ coordinates pair the coefficients of polynomials with points of the moment curve, which separates the
senders that each receiver selects; two further coordinates then give all $k_j$ the same norm.

Give each tile $j$ a real tag $t_j$, the tags distinct, and use moment-curve
separation\cite{donoho2005nonnegative}. For each receiver $i$ put
\[
p_i(t)=\prod_{j\in A(i)}(t-t_j)^2,\qquad \bar p=\frac1T\sum_i p_i .
\]
Row $i$ of $A$ has $K$ ones, so each $p_i$ is monic of degree $2K$; it is also nonnegative. The mean $\bar p$ is then
monic of degree $2K$ as well, and the leading terms cancel in the difference:
\[
\deg(\bar p-p_i)\le 2K-1 .
\]
Let $q_i\in\mathbb R^{2K}$ be the coefficients of $\bar p-p_i$ in $(1,t,\ldots,t^{2K-1})$, so that
\begin{equation}
q_i^\top(1,t,\ldots,t^{2K-1})=\bar p(t)-p_i(t)\qquad\text{for every real } t .
\label{eq:real_coeff}
\end{equation}
Coefficients are linear in the polynomial, and the differences sum to zero,
\[
\sum_i(\bar p-p_i)=T\bar p-\sum_i p_i=0,
\]
so $\sum_i q_i=0$.

Next we evaluate at the tags. As $A$ is zero-diagonal, $j\notin A(j)$, so no factor of $p_j$ vanishes at $t_j$; since
every $p_i$ is nonnegative,
\[
\bar p(t_j)\ge\frac{p_j(t_j)}{T}>0 .
\]
We may therefore put
\[
v_j=\frac{(1,t_j,\ldots,t_j^{2K-1})}{\bar p(t_j)} ,
\]
and equation~\eqref{eq:real_coeff} at $t=t_j$ gives
\[
q_i^\top v_j=\frac{\bar p(t_j)-p_i(t_j)}{\bar p(t_j)}=1-\frac{p_i(t_j)}{\bar p(t_j)} .
\]
The tags are distinct, so $p_i(t_j)=0$ exactly when $j\in A(i)$, and $p_i(t_j)>0$ otherwise. Hence $q_i^\top v_j$
equals $1$ if and only if $j\in A(i)$:
\begin{equation}
q_i^\top v_j=1\quad\text{for } j\in A(i),\qquad q_i^\top v_j<1\quad\text{otherwise}.
\label{eq:real_sep}
\end{equation}
At a selected tag $t_j$, $j\in A(i)$, the polynomial $\bar p-p_i$ takes the value $\bar p(t_j)>0$. It is not the zero
polynomial, so $q_i\ne0$.

Centring keeps this separation. Put $\bar v=\frac1T\sum_j v_j$ and $k'_j=v_j-\bar v$, so that $\sum_j k'_j=0$. Then
\[
q_i^\top k'_j=q_i^\top v_j-q_i^\top\bar v ,
\]
which shifts every entry of row $i$ by the same amount $q_i^\top\bar v$. With $c_i=1-q_i^\top\bar v$,
equation~\eqref{eq:real_sep} becomes
\begin{equation}
q_i^\top k'_j=c_i\quad\text{for } j\in A(i),\qquad q_i^\top k'_j<c_i\quad\text{otherwise}.
\label{eq:real_sepc}
\end{equation}

It remains to give the $k'_j$ one norm without changing their products with the $q_i$. Choose $R>0$ with
\[
R^2\ge2\max_j\|k'_j\|^2 ,
\]
and put $\sigma_j=(R^2-\|k'_j\|^2)^{1/2}$. Each $\sigma_j$ then lies in $[R/\sqrt2,\,R]$:
\[
\frac{R^2}{2}\le R^2-\max_l\|k'_l\|^2\le\sigma_j^2\le R^2 .
\]
For $T\ge3$, every $\sigma_j$ is smaller than the sum of the others,
\[
\sum_{l\ne j}\sigma_l\ge(T-1)\,\frac{R}{\sqrt2}\ge\sqrt2\,R>R\ge\sigma_j ,
\]
so a closed planar polygon has sides $\sigma_j$. Its edge vectors $z_j\in\mathbb R^2$ satisfy
\[
\sum_j z_j=0,\qquad \|z_j\|=\sigma_j .
\]

Finally put $k_j=(k'_j,z_j)$ and append two zeros to $q_i$, so that both lie in $\mathbb R^{2K+2}$. The appended zeros
leave every product $q_i^\top k_j=q_i^\top k'_j$ unchanged, so equation~\eqref{eq:real_sepc} holds for the $k_j$ with
the same $c_i$. The sums vanish block by block, and the norms are equal:
\[
\sum_j k_j=\Bigl(\sum_j k'_j,\ \sum_j z_j\Bigr)=0,\qquad
\|k_j\|^2=\|k'_j\|^2+\|z_j\|^2=\|k'_j\|^2+\sigma_j^2=R^2 .
\]
With $\sum_i q_i=0$ and $q_i\ne0$ from above, these are the claims of the lemma.
\end{proofblock}

\begin{proofblock}[Realizing paired masks]
Lemma~\ref{lemma:ka_realisation} realizes any pair of masks in one head pair. Let $A$ be a non-self row-$K$ mask and
$B$ a non-self column-$K$ mask, and apply Lemma~\ref{lemma:ka_realisation} to each:
\begin{align*}
\text{to } A\colon&\quad \text{pull receivers } q_i,\ \text{senders } k_j;\\
\text{to } B^\top\colon&\quad \text{push senders } q_j,\ \text{receivers } k_i.
\end{align*}
In each branch the ranked family, the vectors $k$, has equal norms. All families are centred, so uniform tile centring
leaves them unchanged. RMS normalization with fixed $\epsilon\ge0$ scales the ranked family by one positive factor and
every other vector by its own. For each receiver in pull, and each sender in push, the scores are therefore one
positive multiple of the products $q^\top k$ of Lemma~\ref{lemma:ka_realisation}, so row-top-$K$ (pull) and
column-top-$K$ (push) return $A$ and $B$ with strict gaps, unaffected by tie-breaking or self-exclusion.

Zero-padding embeds the families in any $d_k\ge2K+2$. Concatenating the four families, two per branch, gives the
input width
\[
4\,(2K+2)=8K+8\le C ,
\]
and the independent branch projections select blocks. Tile-constant features make normalized pooling exact for any
positive token counts. With constant positional terms and gains, and Derf parameters fixed at the given flow times, a
common positive scale places the finite features in the open range of Derf, which is inverted coordinatewise; positive
scaling preserves rankings.
\end{proofblock}

\begin{lemma}[Forced tree]\label{lemma:ka_forced}
Let $\mathcal T$ be a spanning tree of the graph whose edges are all $T^2$ pairs between the $T$ receivers and
the $T$ senders. Give pair $(i,j)$ cost $0$ if $(i,j)\in\mathcal T$ and $1$ otherwise, and let $a_i$ and $b_j$ be
the numbers of pairs of $\mathcal T$ at receiver $i$ and at sender $j$. Then $\mathcal V=0$, and
$\mathcal V(\mathcal G)\ge1$ for every set $\mathcal G$ of pairs that misses a pair of $\mathcal T$.
\end{lemma}

\begin{proofblock}
Unit mass on each pair of $\mathcal T$ is a plan with these masses, and it costs $0$. The masses are positive, as the
tree spans every receiver and sender, with equal totals
\[
\sum_i a_i=\sum_j b_j=|\mathcal T| .
\]
Costs are nonnegative, so $\mathcal V=0$.

Now let $(i_0,j_0)\in\mathcal T\setminus\mathcal G$, and let $\mathcal X$ be the vertex set of the part of
$\mathcal T\setminus\{(i_0,j_0)\}$ that contains receiver $i_0$, so that sender $j_0$ lies outside $\mathcal X$. Every
other pair of $\mathcal T$ has both ends in $\mathcal X$ or neither, and adds one unit to both sums below or to neither;
$(i_0,j_0)$ adds one unit to the first only. The receivers in $\mathcal X$ therefore carry one unit more mass than the
senders in $\mathcal X$:
\[
\sum_{i\in\mathcal X}a_i-\sum_{j\in\mathcal X}b_j=1 .
\]
A plan $\Phi$ that $\mathcal G$ contains therefore moves at least one unit from senders outside $\mathcal X$ to
receivers in $\mathcal X$. Split the mass of these receivers by where it comes from, and bound the part from senders in
$\mathcal X$ by their total mass:
\begin{align*}
\sum_{i\in\mathcal X}a_i
&=\sum_{i\in\mathcal X}\sum_{j\in\mathcal X}\Phi_{ij}+\sum_{i\in\mathcal X}\sum_{j\notin\mathcal X}\Phi_{ij}\\
&\le\sum_{j\in\mathcal X}b_j+\sum_{i\in\mathcal X}\sum_{j\notin\mathcal X}\Phi_{ij} ,
\end{align*}
and with the balance above the last double sum is at least $1$. Only $(i_0,j_0)$, which $\mathcal G$ misses, joins
them within $\mathcal T$; every pair of $\mathcal G$ from a sender outside $\mathcal X$ to a receiver in $\mathcal X$
therefore costs $1$, so that unit moves at cost $1$. Hence $\mathcal V(\mathcal G)\ge1$.
\end{proofblock}

\begin{proofblock}[Proof of (i)]
We take optimal plans whose supports are forests, enlarge the union of these forests to $K$ edge-disjoint spanning
trees within $E$, split each tree into pull and push selections, and realize the selections with
Lemma~\ref{lemma:ka_realisation}.

Optimal vertices have forest support. Choose an optimal vertex $\Phi_\ell^*$ of each compact polytope $\mathcal U_\ell$.
Its bipartite support is a forest, as for every vertex of a transportation polytope\cite{peyre2019ot}. The union of
their supports within $E$,
\[
F=\bigcup_{\ell=1}^{K}\bigl(\operatorname{supp}\Phi_\ell^*\cap E\bigr),
\]
partitions into at most $K$ forests, assigning each repeated pair to just one.

We now enlarge $F$ to $K$ edge-disjoint spanning trees within $E$. Sets of pairs that split into at most $K$ forests
are the independent sets of the union of $K$ graphic matroids\cite{edmonds1965union} on $E$. A forest on the $T$
receivers and $T$ senders has at most $2T-1$ pairs, so this union has rank at most $K(2T-1)$. We show that the rank
equals $K(2T-1)$ by exhibiting $K$ disjoint spanning trees within $E$, and then extend $F$ to a basis.

Label the tiles $0,1,\ldots,T-1$ and read labels modulo $T$. For each shift $s$, collect the pairs whose sender lies
$s$ places after the receiver:
\[
M_s=\{(i,j):j-i\equiv s\pmod T\},\qquad 1\le s\le T-1 .
\]
Each receiver and each sender lies in exactly one pair of $M_s$, and $s\not\equiv0$ excludes the self pairs, so $M_s$
is a perfect matching within $E$. In the receiver-by-sender array of Supplementary Fig.~\ref{suppfig:shifts}b, $M_s$ is
the cyclic diagonal of cells marked $s$.

Two consecutive shifts form a Hamiltonian cycle. Let $2r\le T-1$. From receiver $i$, the pair of $M_{2r}$ leads to
sender $i+2r$, and that sender's pair in $M_{2r-1}$ leads back to receiver $i+1$:
\[
\text{receiver } i\ \xrightarrow{\;M_{2r}\;}\ \text{sender } i+2r\ \xrightarrow{\;M_{2r-1}\;}\
\text{receiver } i+1 .
\]
Starting from receiver $0$, these moves pass through the receivers $0,1,\ldots,T-1$ and the senders
$2r,2r+1,\ldots,2r+T-1$, each once, and close at receiver $0$. So $M_{2r-1}\cup M_{2r}$ is a Hamiltonian cycle, drawn
for $r=1$ in Supplementary Fig.~\ref{suppfig:shifts}a. Deleting its pair $(0,2r)$ leaves a Hamiltonian path,
\[
Q_r=\bigl(M_{2r-1}\cup M_{2r}\bigr)\setminus\{(0,2r)\},
\]
a spanning tree of $2T-1$ pairs within $E$. Different $r$ use different shifts, so
\[
Q_r\cap Q_{r'}=\varnothing\qquad(r\ne r').
\]
For $T\ge2K+1$, the condition $2r\le T-1$ holds for $r=1,\ldots,K$. The paths $Q_1,\ldots,Q_K$ then exist, and
together they have $K(2T-1)$ pairs, which attains the rank bound.

\par\noindent\begin{minipage}{\textwidth}
\suppfig{Modular-shift matchings and the spanning trees $\boldsymbol{Q_r}$.}\label{suppfig:shifts}
{\centering
\begin{tikzpicture}[x=1.05cm,y=1.55cm,
    every node/.style={font=\sffamily\small},
    vtx/.style={circle,fill=black,inner sep=1.5pt},
    mtwo/.style={line width=1.1pt,black},
    mone/.style={line width=1.1pt,black!45}]
  \node[font=\sffamily\bfseries] at (-1.05,1.78) {a};
  \foreach \i in {0,...,4}{
    \node[vtx] (r\i) at (\i,1) {};
    \node[above=2pt] at (r\i) {\i};
    \pgfmathtruncatemacro{\j}{mod(\i+2,5)}
    \node[vtx] (s\i) at (\i,0) {};
    \node[below=2pt] at (s\i) {\j};
  }
  \node[anchor=west] at (4.7,1) {receivers};
  \node[anchor=west] at (4.7,0) {senders};
  \draw[mtwo,dashed] (r0) -- (s0);
  \foreach \i in {1,...,4}{ \draw[mtwo] (r\i) -- (s\i); }
  \foreach \i/\k in {0/1,1/2,2/3,3/4}{ \draw[mone] (s\i) -- (r\k); }
  \draw[mone] (s4) to[out=-25,in=0] (2,-0.78) to[out=180,in=-90] (-0.72,0.35)
              to[out=90,in=180] (r0);
  \begin{scope}[shift={(7.9,1.55)},x=0.52cm,y=0.52cm]
    \node[font=\sffamily\bfseries] at (-2.6,0.69) {b};
    \foreach \i in {0,...,4}{
      \node at (-0.5,-\i-0.5) {\i};
      \node at (\i+0.5,0.45) {\i};
      \foreach \j in {0,...,4}{
        \pgfmathtruncatemacro{\s}{mod(\j-\i+5,5)}
        \ifnum\s=0
          \draw[black!35] (\j,-\i) rectangle ++(1,-1);
        \else
          \pgfmathtruncatemacro{\del}{(\i==0)&&((\j==2)||(\j==4))}
          \ifnum\del=1
            \draw[black!70,dashed] (\j+0.04,-\i-0.04) rectangle ++(0.92,-0.92);
          \else
            \ifnum\s<3
              \fill[black!42] (\j,-\i) rectangle ++(1,-1);
            \else
              \fill[black!12] (\j,-\i) rectangle ++(1,-1);
            \fi
          \fi
          \node[font=\sffamily\footnotesize] at (\j+0.5,-\i-0.5) {\s};
        \fi
      }
    }
    \draw[black!60] (0,0) rectangle (5,-5);
    \node[rotate=90] at (-1.45,-2.5) {receiver \textit{i}};
    \node at (2.5,1.15) {sender \textit{j}};
  \end{scope}
\end{tikzpicture}\par}\vspace{1mm}
\par\noindent{\footnotesize
\textbf{a},~The Hamiltonian cycle $M_1\cup M_2$: each sender is drawn under the receiver that $M_2$ pairs it with
(black), and grey pairs belong to $M_1$. Deleting the dashed pair $(0,2)$ leaves the path $Q_1$.
\textbf{b},~The pairs $(i,j)$ as a receiver-by-sender array, each cell marked with its shift $j-i \bmod T$. $Q_1$
(dark) and $Q_2$ (light) are disjoint spanning trees; the diagonal holds the self pairs, and the dashed cells are the
deleted pairs $(0,2)$ and $(0,4)$.\par}
\end{minipage}\par\medskip

Matroid basis extension now enlarges $F$ to a basis $\mathcal H$ of the union matroid, the disjoint union of $K$
forests $\mathcal T_r$:
\[
F\subseteq\mathcal H=\mathcal T_1\cup\cdots\cup\mathcal T_K,\qquad |\mathcal H|=K(2T-1).
\]
Each forest has at most $2T-1$ pairs, and the $K$ forests have $K(2T-1)$ pairs in total, so every $\mathcal T_r$ has
exactly $2T-1$ pairs and is a spanning tree.

Each optimal vertex $\Phi^*_\ell$ moves mass only along $I\cup F\subseteq I\cup\mathcal H$, so $I\cup\mathcal H$
contains it, and restricting the pairs cannot lower the minimum:
\[
\mathcal V_\ell\le\mathcal V_\ell(I\cup\mathcal H)\le\langle D_\ell,\Phi^*_\ell\rangle=\mathcal V_\ell .
\]
Thus $\mathcal H$ holds the collection.

The trees are now split into pull and push selections. Root each $\mathcal T_r$. Each non-root vertex selects its
parent pair, through pull if it is a receiver and through push if it is a sender, and the root selects one of its
pairs. In each tree every receiver then selects one pair through pull and every sender one through push. The
edge-disjoint trees therefore give a row-$K$ mask $A$ and a column-$K$ mask $B$, both non-self as $\mathcal H\subseteq E$,
with
\[
A\cup B=\mathcal H.
\]

Finally, the construction for paired masks above realizes $A,B$ with the stated widths, in exact arithmetic. Thus
\[
\Pi^\circ=A^{\mathrm{pull}}\cup A^{\mathrm{push}}=A\cup B=\mathcal H,
\]
and one head pair has $K(2T-1)$ selected pairs, which hold the collection.
\end{proofblock}

\begin{proofblock}[Proof of (ii)]
Put $d=mK$, the most non-self pairs that the $m$ heads select in one row, when they select at the receiver, or in one
column, when they select at the sender. Fix a tile $i$ and let $\mathcal D$ be the pairs in row $i$ and in column $i$.
Receiver $i$ is joined to every sender and sender $i$ to every receiver, and the two stars share $(i,i)$, so
$\mathcal D$ is a spanning tree centred on $(i,i)$ with
\[
|\mathcal D|=T+T-1=2T-1
\]
pairs. Give it the costs and masses of Lemma~\ref{lemma:ka_forced}.

The heads place at most $d$ non-self pairs in row $i$, or in column $i$, of the $T-1$ that $\mathcal D$ has there, so
at least $(T-1)-d$ pairs of $\mathcal D$ are missing from the self pairs and the selected pairs. Each missing pair is the only pair of $\mathcal D$ at
its other end, which carries one unit of mass, so that end's unit moves off $\mathcal D$, at cost $1$. Every held plan
therefore moves at least $T-1-mK$ units at unit cost.

For $K+1\le T\le d+1$, the heads can select all of $E$. Give them disjoint blocks of at most $K$ of the shifts
$1,\ldots,T-1$, each padded to $K$ shifts, which $T-1\ge K$ allows. Through the matchings $M_s$ of the proof of (i),
whose union over $s=1,\ldots,T-1$ is $E$, the heads then reach every other tile, and so hold every plan.
\end{proofblock}

\begin{proposition}[Least budget]\label{prop:ka_budget}
(a) If $T\ge2K+1$, every collection of $K$ transport problems as in Theorem~\ref{thm:ka_support} is held by some
$K(2T-1)$ non-self pairs, and some collection by no fewer.\\
(b) If $T\ge2K+2$, some collection of $K+1$ such problems is held by no set of $2TK$ non-self pairs, so one head
pair can hold every collection of $K$ but not every collection of $K+1$.
\end{proposition}

\begin{proofblock}
(a) Let $T\ge2K+1$, and recall the disjoint paths $Q_1,\ldots,Q_K$ built in the proof of Theorem~\ref{thm:ka_support}(i).
Give problem $\ell$ the costs and masses of Lemma~\ref{lemma:ka_forced} for $Q_\ell$, $\ell=1,\ldots,K$. By
Lemma~\ref{lemma:ka_forced}, a set $\mathcal G$ of non-self pairs that holds them contains each $Q_\ell$, which has no
self pair, so
\[
|\mathcal G|\ge|Q_1\cup\cdots\cup Q_K|=\sum_\ell|Q_\ell|=K(2T-1);
\]
the head pair of (i) attains this for every collection.

(b) Let $T\ge2K+2$. Then $2K+1\le T-1$, so the matching $M_{2K+1}$ also lies in $E$, and it meets no $Q_\ell$, whose
shifts are at most $2K$. Extend it by pairs of $Q_1$ to a spanning tree within $E$ and add its problem. Every such
$\mathcal G$ now contains $Q_1,\ldots,Q_K$ and $M_{2K+1}$, so
\[
|\mathcal G|\ge K(2T-1)+T=2TK+(T-K)>2TK .
\]
A head pair selects at most $2TK$ pairs, by equation~\eqref{eq:ka_support_count}, so by
Theorem~\ref{thm:ka_support}(i) one head pair can hold every collection of $K$ but not every collection of $K+1$.

For $h$ head pairs with independent scores, which exchange tokens over the self pairs and the union of their
selected pairs, the same arguments with $hK$ for $K$ give three statements. For $T\ge2hK+1$, $d_k\ge2K+2$ and
$C\ge h(8K+8)$, every collection of $hK$ problems is held by $hK(2T-1)$ distinct selected pairs, $K$ of the $hK$
spanning trees going to each head pair on its own $8K+8$ input features. For $T\ge2hK+2$, some collection of $hK+1$
needs more than $2ThK$ pairs, so the $h$ head pairs hold exactly $hK$. For $T\le2hK+1$, they can select all of $E$.
\end{proofblock}

\begin{proposition}[Shared scores]\label{prop:ka_shared}
Let pull and push select by one score matrix, with ties broken by one order of the pairs. Then every collection
of $\lfloor(K+1)/2\rfloor$ transport problems as in Theorem~\ref{thm:ka_support} is held by the pairs selected under
some score matrix, and, for $T\ge K+2$, some collection of $\lfloor(K+1)/2\rfloor+1$ such problems by those of none.
\end{proposition}

\begin{proofblock}
We first describe the sets that one shared order selects whole, and then count pairs in unions of forests.

One score matrix and one tie order rank the pairs of $E$ by one strict order for both branches. A set
$\mathcal W\subseteq E$ is then selected whole for some order if and only if
\begin{equation}
\begin{gathered}
\text{every nonempty } \mathcal W'\subseteq\mathcal W\text{ has a pair with an end, receiver or sender,}\\
\text{that holds at most $K$ pairs of } \mathcal W' .
\end{gathered}
\label{eq:shared_peel}
\end{equation}
Suppose first that some $\mathcal W'$ has no such pair. Its lowest-ranked pair $(i,j)$ is then outranked by $K$ pairs
of $\mathcal W'$ in row $i$ and by $K$ in column $j$, so neither branch selects it, and $\mathcal W$ is not selected
whole.

Conversely, suppose that condition~\eqref{eq:shared_peel} holds. Remove pairs of $\mathcal W$ one at a time, each at an
end holding at most $K$ remaining pairs, which the condition allows at every stage, and rank them in the order of
removal, the first lowest, above all of $E\setminus\mathcal W$. At that end, at most $K-1$ pairs then rank above each
pair, which is therefore among that end's top $K$; every pair of $\mathcal W$ is selected.

The non-self supports of optimal vertices of $n$ problems are forests. A nonempty $\mathcal W'$ within their union,
meeting a set $\mathcal X$ of receivers and senders, therefore has at most $n(|\mathcal X|-1)$ pairs. Counting the ends
of these pairs, the members of $\mathcal X$ hold
\[
2|\mathcal W'|\le2n\,(|\mathcal X|-1)<2n\,|\mathcal X|
\]
pairs in total, so some member of $\mathcal X$ holds at most $2n-1$ of these pairs, and for
$n=\lfloor(K+1)/2\rfloor$
\[
2n-1=2\Bigl\lfloor\frac{K+1}{2}\Bigr\rfloor-1\le K .
\]
Condition~\eqref{eq:shared_peel} thus holds for the union, which proves the first part.

For the second, let $T\ge K+2$, so that every shift up to $K+1$ is at most $T-1$, and put
\[
\mathcal C=M_1\cup\cdots\cup M_{K+1}\subseteq E ,
\]
the cells marked $1$ to $K+1$ in the array of Supplementary Fig.~\ref{suppfig:shifts}b. Each receiver and sender holds
$K+1$ pairs of $\mathcal C$, so $\mathcal W'=\mathcal C$ violates condition~\eqref{eq:shared_peel} and no shared order
selects all of $\mathcal C$. Yet $\mathcal C$ splits into $\lfloor(K+1)/2\rfloor+1$ forests: the paths $Q_r$ for
$r\le\lfloor(K+1)/2\rfloor$, and the star of their deleted pairs $(0,2r)$, joined with $M_{K+1}$ when $K$ is even. In
this union no receiver other than $0$ holds two or more pairs, whereas a cycle passes through at least two receivers
that do. Extend each forest by pairs of $Q_1$ to a spanning tree within $E$ and give it the problem of
Lemma~\ref{lemma:ka_forced}. By Lemma~\ref{lemma:ka_forced}, self and selected pairs that contain optimal plans of all these
problems contain every tree, hence $\mathcal C$, which no shared order selects.
\end{proofblock}

\noindent\emph{Remark.} For $T\ge\max(K+1,3)$, $d_k\ge2K+2$ and $C\ge8K+8$, a set $\mathcal W$ of non-self pairs lies
within the selected pairs of one KA head pair, for some KA inputs and parameters, if and only if every set
$\mathcal X$ of receivers and senders is joined by at most $K|\mathcal X|$ pairs of $\mathcal W$. Suppose first that
$\mathcal W$ lies within the selected pairs. Charge each selected pair to the receiver that pulls it or the sender
that pushes it, one of them if both do; each is charged at most $K$. The pairs of $\mathcal W$ that join members of
$\mathcal X$ are charged to members of $\mathcal X$, so there are at most $K|\mathcal X|$ of them, which gives
necessity. Conversely, by a corollary of Hakimi's orientation theorem\cite{hakimi1965}, the count condition allows each pair of
$\mathcal W$ to be charged to one of its ends, at most $K$ to each. Let receivers pull and senders push their charged
pairs, add non-self pairs until every row and column has $K$, which $T-1\ge K$ allows, and realize the two masks as in
the proof of (i).

\medskip
No threshold on $T$ in these statements can be lowered. With $|E|=T(T-1)$:
\begin{alignat*}{2}
T&\le2K\colon&\qquad&|E|<K(2T-1),\ \text{so the count in (i) and Proposition~\ref{prop:ka_budget}(a) fail};\\
T&\le mK+1\colon&&m\text{ heads on one side can select } E,\ \text{so (ii) fails};\\
T&\le2K+1\colon&&|E|\le2TK,\ \text{so Proposition~\ref{prop:ka_budget}(b) fails};\\
T&=K+1\colon&&\text{one score matrix selects } E .
\end{alignat*}
The second line is the end of the proof of (ii). The last concerns Proposition~\ref{prop:ka_shared}: each row and
column then holds only $K$ non-self pairs.
\endgroup

\subsection{S2. Selection statistics and operation counts}
\paragraph{Tiles selected by no receiver.} Over 24 initializations of 2020, the pull branch's selections alone leave 38.0\% of tiles at 1\textdegree\ and 28.2\% at 2\textdegree\ selected by no receiver in a head, averaged over 1,152 and 3,456 head graphs. With the push branch, every tile selects $K$ receivers to send to, by construction.

\paragraph{Maximum-score selection.} The maximum-score binary selection replaces transportation marginals\cite{hitchcock1941} by degree lower bounds, maximizing the scores of one branch and head over $U$ non-self pairs:
\begin{equation}
\begin{aligned}
&\max_{\Gamma\in[0,1]^{T\times T}}\; J(\Gamma) = \sum_{i\neq j} s_{ij}\Gamma_{ij}\\
&\text{subject to}\quad
\Gamma_{ii}=0,\quad \sum_{j}\Gamma_{ij}\ge K,\quad \sum_{i}\Gamma_{ij}\ge K,\quad \sum_{i,j}\Gamma_{ij}=U.
\end{aligned}
\label{eq:lp}
\end{equation}

Unit capacities make this a minimum-cost flow LP with an integral optimum. Its solution at the two-sided rule's
pair count is the optimal selection of Extended Data Fig.~\ref{ed:routing_toy}a.

\paragraph{Cost.} At two operations per multiply--add, the selected-token products, $4(K+1)P^2N'd_k$, and tile scores plus pooling, $4T^2d_k$, sum to $4d_k[(K+1)P^2N'+N'^2/P^4]$ per head and branch. The 32.0\,TFLOP network total is traced. Unshifted tiles give:
\begin{center}
\begin{tabular}{lrr}
\toprule
Quantity & L1 & L2\\
\midrule
Valid points $N$ & 65,160 & 16,380\\
Padded tokens $N'$ & 66,240 & 18,000\\
Tile scores and pooling / selected attention & 2.8\% & 0.2\%\\
Two-branch selected / dense attention & 1.8\% & 12\%\\
\bottomrule
\end{tabular}
\end{center}

The ratios are $N'/[(K+1)P^6]$ and $2(K+1)P^2N'/N^2$, respectively. Dense attention uses the heads of
one KA branch. Replacing L1--L2 attention adds about 120\,TFLOP and removes
3.6\,TFLOP of projections and tile scores from the 32.0\,TFLOP network, giving the reported 4.6-fold cost.

\subsection{S3. Transport task at the 1\textdegree\ level}
The transport task uses \model's 1\textdegree\ grid and tiles ($181\times360$ points, $8\times8$-point
tiles, $T=1{,}035$). Each tile carries a 32-channel payload, $\pm1$ in independent blocks of $4\times4$ points, and two
64-dimensional codes drawn for each task, a destination code and an identity code. Tile $j$ is a source of receiver
$i$ when $j$'s destination code equals $i$'s identity code. At each point of a receiver tile the target is the mean of
the payloads of its $d$ sources at the same position in the tile; the input also gives the position in the tile.
Training tasks draw one of four regimes with equal probability: a permutation ($d=1$); regular groups of 7 or 8
sources and as many receivers; receiver-skewed demand, in which the source counts of the groups follow a truncated
power law with exponent 1.15 up to 96 while each group keeps 7 or 8 receivers; and sender-skewed demand, with the two
roles swapped. The three grouped regimes have 8,245 pairs each, and no tile is its own source.

The network has two residual attention blocks, the second on tiles shifted by half a tile, each followed by a
pointwise feed-forward layer. The two methods differ in the attention operator. \hla\ is \model's
1\textdegree\ operator (two branches of three heads with $d_k=128$, $K=8$). The pull-only variant replaces the push
branch's sender-side selection by a second receiver-side selection and keeps the heads, gates, pooling and $K$.

The error $E$ averages $d\,(\hat y-y)^2$ over points and channels, with prediction $\hat y$ and target $y$, so that the zero output has $E=1$ at every $d$. Each
method is trained from three seeds on freshly drawn tasks with \model's optimizer, and training times sum the
measured times per update of each run on one GPU. Test errors on 128 held-out tasks per regime are taken at the update, among those every 500, with
the lowest validation error.

Up to eight sources, \hla\ and the pull-only variant both stay below 0.1\% of the zero output's error. With 49 or more sources the error of the pull-only variant reaches 58\%, and that of \hla\ 1.1\%, in no more training time.

\subsection{S4. Network architecture and implementation}

The network runs two cycles, encoder~1--decoder~1--encoder~2--decoder~2\cite{ronneberger2015,newell2016} (grids, widths and blocks in Extended Data Table~\ref{edt:cost}a). Resampling is by 4 between L0 and L1
and by 2 elsewhere, with separate parameters per cycle. Additive skip links are encoder~1 to decoder~1
(L1--L3), decoder~1 to encoder~2 (L2--L4), encoder~2 to decoder~2 (L1--L3), and encoder~1 to decoder~2 (L0).
Encoder~1 and decoder~2 have two L0 blocks. Each stage has four L1 and L3 blocks and six L2 blocks;
each cycle has four encoder and four decoder blocks at L4.

L1--L4 blocks use attention and FFN residual connections, each preceded by a Derf layer (Extended Data Fig.~\ref{ed:arch}b). KA adds the within-tile position table before
its Derf layer. L0 uses a $3\times3$ convolutional FFN. L3 and L4 use dense attention over 4,140 and 1,035
tokens, with 15 and 24 heads of dimension 128. Their axial rotary embeddings\cite{su2024} have integer
longitude wavenumbers. L1--L4 FFNs use SwiGLU\cite{shazeer2020} and a depthwise $3\times3$ convolution (Extended Data Fig.~\ref{ed:arch}d)
after expansion\cite{zamir2022}.

Except for $\rho$ (equation~\eqref{eq:nodes}) and the scaling to unit RMS over channels of the two inputs of each resampling attention residual connection, the network uses point-wise Derf\cite{chen2025derf} (Extended Data Fig.~\ref{ed:arch}c) in place of normalization layers, which needs no token or channel statistics:
\begin{equation}
\mathrm{Derf}(\mathbf{h}) = \boldsymbol{\gamma}(r,\tau) \odot
\mathrm{erf}(\omega\,\mathbf{h} + \delta) + \boldsymbol{\beta}(r,\tau),
\label{eq:derf}
\end{equation}
with learned per-layer scalars $\omega>0$, $\delta$. A two-layer perceptron (64 hidden units) maps sinusoidal features of the MeanFlow times $\tau$ and $\tau-r$ (the flow-time features) to channel-wise $\boldsymbol\gamma$, $\boldsymbol\beta$, initially 1 and 0. These features also condition \hla's score gates and a rotary phase: every attention layer rotates each head's queries by $2\tanh(\nu)$. Here $\nu$ is a zero-initialized learned affine function of token and flow-time features, per level and, in \hla, per branch.

Encoder~1 receives $1\times1$ projections of the four input states, static fields and forcings at every level,
area-averaged to its grid; L0 and L1 also receive $\mathbf z_\tau$. A latitude embedding is added at each
level's feature entry points. The output head applies a latitude-row/channel gain and grid-point scale and
shift from orography, sine of latitude and land--sea mask, then projects to 75 variables. Poleward of
60\textdegree, its longitude filter passes wavenumbers through $k_c=\max(1,1{,}440\cos\phi)$ and removes
those above $1.5k_c$, with a $\cos^2$ taper between. The five land/ocean-domain fields are not filtered.

\paragraph{Remaining parameters.} L0--L2 FFN hidden activations have latitude-row/channel scale and shift.
SwiGLU width is $8/3$ of the level width; the L0 FFN expands 96 channels to 192. Convolutions and resampling wrap in longitude; across a pole they
learn a per-channel mix of the edge rows and the rows continued over the pole with a 180\textdegree\
longitude rotation. Resampling applies a 12-tap Kaiser-windowed sinc at the coarser Nyquist cut-off
(twice between L0 and L1), a $1\times1$ projection, then an attention residual over eight fixed parts of
each matching patch in a $3\times3$ neighbourhood.

\paragraph{Initialization.} Dense weights use zero-mean Gaussian initialization with variance $1/f_{\rm in}$.
Residual output projections have standard deviation $0.7/\sqrt{n f_{\rm in}}$, where $n$ counts residual
writes in the corresponding encoder or decoder program at that level, including its incoming resampler.
Positional tables have standard deviation 0.02, and depthwise kernels start as spatial identities. Projections
for the two oldest history states and the forcing fields, the final output head and flow-time phase
coefficients start at zero.
Derf input scales are calibrated by fixed-point iteration on the freshly initialized network to a median
activation RMS of 0.70711 per layer group at $(r,\tau)=(0.2,0.5)$.

\subsection{S5. Data, conditioning and noise prior}
\paragraph{Data.} ERA5's equiangular grid includes both poles (variables and transforms in Extended Data Table~\ref{edt:vars}b; sampling and years in Methods). Predicted 6\,h residuals are zero for the five slowly varying auxiliary fields outside their land/ocean domains, for sea-ice cover outside the ice-possible mask, for sea-surface temperature where ice cover is at least 0.5 and for snow depth where it is at least 1\,m water equivalent.

\paragraph{Conditioning.} The 36 static fields are orography and five fields of its sub-grid structure (standard deviation, slope, anisotropy, and twice its angle as sine and cosine); the land--sea mask; lake cover; high- and low-vegetation cover; indicator fields of vegetation types (17) and soil types (8); and an ice-possible mask, the grid points where sea-ice cover was non-zero at any time in 1979--2019.

Of the 13 forcing fields, eight are solar: the top-of-atmosphere insolation over the preceding and coming 6\,h, the cosine of the solar zenith angle now and 6\,h ahead, and the local hour angle and the phase of the year, each as its sine and cosine. The other five, the surface forcings, are taken from 00\,UTC of a source day: the anomaly of sea-surface temperature from its 1979--2019 mean for that day of the year, sea-ice cover and top-layer soil water, with the age of these fields and an indicator that the age is under one day. In training, the source day is that of the current state with probability 0.5 and otherwise 1--16 days earlier, drawn uniformly. In a forecast it is the day of initialization (Methods).

\paragraph{Noise.} Each step starts from a noise prior fitted to ERA5 6\,h residuals of 1998--2019 after removing means for each hour of day and season (Extended Data Fig.~\ref{ed:prior}). White noise $\boldsymbol\xi$ is coloured in spherical harmonics with order-dependent degree spectra (anisotropy) per variable and cross-variable correlations per degree band. Degrees up to 63 are further mixed across variables per latitude row, separately over land and sea. Wind receives a regression on the same-level geopotential gradient (mean sea-level pressure for 10\,m wind), with latitude-, surface-type- and degree-dependent coefficients. Each latitude row is normalized, then scaled by the local residual standard deviation after the same mean removal, with amplitudes normalized to unit grid-mean square. The five slowly varying auxiliary fields have zero noise and zero $\mathbf z_\tau$ input; their changes depend on conditioning alone.

\subsection{S6. Training objectives and optimization}

\paragraph{First stage: MeanFlow.} Define $\mathbf z_0=(\mathbf x_{t_0+6\mathrm h}-\mathbf x_{t_0})/
\boldsymbol\sigma_\Delta$, $\mathbf z_\tau=(1-\tau)\mathbf z_0+\tau\boldsymbol\varepsilon$ and
$\mathbf v=\boldsymbol\varepsilon-\mathbf z_0$. Half of each batch uses $r=\tau$, $\tau\sim U[0,1]$ and
target $\mathbf v$; the other half samples $(r,\tau)$ uniformly on $0\le r<\tau\le1$ and uses the MeanFlow target\cite{lipman2023,geng2025}, computed with a Jacobian--vector product and without gradient through the target. For each state and variable, the area-weighted squared error is divided by its own value plus $10^{-4}$, held constant in the gradient, as in MeanFlow's adaptive loss weighting. Variables are then weighted as in Extended Data Table~\ref{edt:hyp}c.

\paragraph{Noisy adapter.} In the last block of decoder~1 at L3 and at L4, the attention output projection $W$ becomes
\begin{equation}
\mathbf{o} = W\mathbf{h} + B\bigl(\boldsymbol{\eta}\odot A\mathbf{h}\bigr),
\label{eq:noisylora}
\end{equation}
a rank-16 adapter\cite{hu2022}. The standard-normal code $\boldsymbol\eta\in\mathbb R^{16}$ is fresh for every member and block but shared by all tokens, perturbing each member coherently at the two coarsest levels. $A$ has the standard fan-in initialization. $B$ is a Gaussian draw scaled so that $\lVert BA\rVert_{\mathrm{F}} = 0.1\,\lVert W\rVert_{\mathrm{F}}$, and both factors are trained with the network. Members draw fresh codes at each forecast step, in training and in forecasts. The frozen copy of SF-CRPS runs without the adapter.

\paragraph{Field term.} The Laplacian pyramid\cite{burt1983} splits each field in $\boldsymbol\sigma_\Delta$ units into four detail bands and a 4\textdegree\ remainder, with exact reconstruction. Each band's area-weighted CRPS receives its band weight (Extended Data Table~\ref{edt:hyp}c) and a fixed normalizer per band and variable: the variable's grid-point CRPS divided by the band's CRPS, measured once on two-step MeanFlow forecasts at stage entry. The band sum is divided by the members' mean absolute error averaged over the globe (the first term of equation~\eqref{eq:afcrps}) plus $10^{-2}$, one stop-gradient number per initialization, variable and lead time. The five noiseless auxiliary fields use only the first term, over their own domains, with an offset of 1 instead of $10^{-2}$ because their 6\,h residuals are often zero.

\paragraph{Clipping and precipitation penalty.} Feedback states pass through a fixed clip in normalized channel units; gradients follow its slope. For soft upper sides, the identity threshold is $b=M+0.25|M|$ and the saturation limit $B=b+0.5|b|$, where $M$ is the channel's maximum over all 6-hourly training states of 1979--2019. A tanh joins the identity to this limit with continuous value and slope; lower soft sides are symmetric. For 6\,h precipitation, the upper threshold and limit are $M+0.25$ and $M+0.75$. Variables with physical bounds are instead clamped to their physical ranges. Where the ERA5 training data lie outside a physical range, as for the known small negative values of specific humidity, the clamp is set at the extreme of the training data. A penalty $c\sum_p\bigl[(D_p+1)^3-1\bigr]$, with $c = 1.1\times10^{-4}$ and $D_p$ the excess of the unclipped 6\,h precipitation at grid point $p$ over its 1979--2019 maximum there (normalized units; zero below it), averaged over initializations, members and lead times, discourages precipitation beyond the record.

\paragraph{Optimizer.} Under schedule-free optimization\cite{defazio2024}, Muown\cite{lion2026muown} updates row directions with Muon\cite{jordan2024muon} and row norms with Adam\cite{kingma2015}. Muown acts on
non-zero-initialized fully connected matrices except the adapters, FFN output projections and resampling
attention output projections. Adam updates these exceptions, zero-initialized matrices, convolution kernels
and learned tables. Hyperparameters are in
Extended Data Table~\ref{edt:hyp}c.

\subsection{S7. Training budgets and inference cost}
Training budgets use U-Cast's conversion factors\cite{cachay2026} in V100 units: 1~A100 = 2, 1~H100 = 4, 1~TPUv5 = 6~V100 and 1~H200 = 2.5~A100 = 5~V100. U-Cast follows the hardware-comparison framework of ArchesWeatherGen\cite{couairon2026}. \model's 449.3 H200-hours are 93.6~V100-days.

GenCast: 160 TPUv5-days (32 devices for 5 days\cite{price2025}, as listed by U-Cast) at 6~V100 per device, 960~V100-days.

SFNO-BVMC\cite{mahesh2025}: the 29 checkpoints of its ensemble, each trained for 16~h on 256 A100 GPUs, i.e.\ 4,949 A100-days, 9,899~V100-days.

FourCastNet~3\cite{bonev2025}: 1,024 H100 GPUs for 78~h, 512 A100 GPUs for 15~h and 256 H100 GPUs for 8~h, i.e.\ 3,413 H100-days and 320 A100-days, 14,293 V100-days.

U-Cast and U-Cast (DeepEns) enter at their published 8.2 and 11.8 H200-days, GraphCast and FuXi at the 2,688 and 52 V100-days of Table~1 of ArchesWeatherGen. ArchesWeatherGen enters at its published 45 V100-days, and Stormer and NeuralGCM ENS at the 256 and 7,680 V100-days of the same table.

For Fig.~\ref{fig:efficiency}c, the training cost is divided by the number of grid points of each model: $721\times1{,}440$ at 0.25\textdegree, $121\times240$ at 1.5\textdegree\ and $128\times256$ at 1.4\textdegree.

\paragraph{Inference cost.} Time/member is 60 times the median step, excluding the first. Memory is the peak allocated memory. \model's network uses float32 with JAX-default
TF32 products; FourCastNet~3 uses bfloat16 mixed precision (the same timing as float32/TF32).
On one RTX~5090, a member of \model\ takes 41\,s and 19.0\,GiB. GenCast is run with its released 0.25\textdegree\ checkpoint and sampler in float32 with JAX-default TF32 products, with the attention settings of its GPU guide
(\url{https://github.com/google-deepmind/weathernext/blob/89c4b2a77a1c57b328b909c575550fd2e5aadc9c/docs/weathernext1_gen/cloud_vm_setup.md}).
One member takes 30 steps of 12\,h, and its time is 30 times the median step, excluding the first.

\subsection{S8. Forecast evaluation}
Like those of \model, GenCast's members start from unperturbed ERA5 states (the initial time and 12\,h earlier) and differ only through sampling noise\cite{price2025}. Initializations run from 1 January 00\,UTC to 31 December 12\,UTC 2020, the initialization times of IFS ENS and of GenCast in WeatherBench~2.

\paragraph{Scores.} We use WeatherBench~2's \texttt{crps}, \texttt{unbiased\_\allowbreak mean\_\allowbreak rmse} and reciprocal
\texttt{unbiased\_\allowbreak spread\_\allowbreak skill}\cite{rasp2024,ferro2014,fortin2014}. Member wind speeds use both components; 24\,h precipitation
sums the four 6\,h accumulations ending at the valid time. The released scoring code evaluates the 0.25\textdegree\ members as they are generated, so the full fields do not have to be stored. On the same GenCast fields (50 and all 56 members) it agrees with WeatherBench~2 to $3.1\times10^{-6}$ in relative CRPS and $1.1\times10^{-6}$ in ensemble-mean RMSE. Other 0.25\textdegree\ scores are the 2020 WeatherBench~2 result files.

At 1.5\textdegree, each of the 50 members was regridded to the $121\times240$ WeatherBench~2 grid with its first-order conservative regridder\cite{jones1999}, wind speeds from each member's 0.25\textdegree\ speeds, as WeatherBench~2 regrids those of ERA5 and GenCast. These regridded members are the released forecasts and WeatherBench~2 scored them. The scores of GenCast and IFS ENS on the same grid are taken from its result files. The result files are produced with WeatherBench-X (\url{https://github.com/google-research/weatherbenchX}).

\subsection{S9. Power spectra and statistical uncertainty}
Spectra use members 0--7 of \model, GenCast and IFS ENS at 0.25\textdegree\ (the latter two from WeatherBench~2) and ERA5 at matching valid times. The 24 initializations are the 2nd and 16th of each month, at 00 and 12\,UTC respectively in odd months and 12 and 00\,UTC in even months. Exact spherical-harmonic transforms on the $721\times1{,}440$ grid give power at total wavenumbers $\ell=1$--719. Pole rows are replaced by their zonal mean (zonal wavenumber-1 component for wind), giving one scalar value or wind vector per pole. The 10\,m wind speed is computed at each grid point from the two components. Pressure-level temperature, geopotential and wind, represented by ERA5 at TL639, are shown to $\ell=638$. Power per degree is divided by $4\pi$ and averaged over members and initializations; its sum is the global area-weighted mean square. The agreement limit of Methods is reported as the wavelength $2\pi R/\ell$ of the largest agreeing degree $\ell$, with $R$ the Earth's radius.

\paragraph{Confidence intervals.} The widening factor of Methods is calibrated on 16 comparisons with IFS ENS at 24\,h, for which the scores of IFS ENS are available for each initialization: CRPS and RMSE for geopotential and temperature at 500, 700 and 850\,hPa, 2\,m temperature and mean sea-level pressure. For each, we compute the interval with both systems resampled together and with IFS ENS held at its published average. The factor is the median ratio of the first width to the second, 1.09. The same bootstrap gives 95\% bands for \model's scores in Extended Data Fig.~\ref{ed:curves}. Skill intervals hold IFS ENS fixed and use the same widening; spread--skill intervals are unwidened.

\begingroup
\renewcommand{\bibsection}{\subsection*{Supplementary References}}

\endgroup

\end{document}